\documentclass[twoside,11pt]{article}
\usepackage{hyperref}       % hyperlinks
\usepackage{jair, theapa, rawfonts}

\usepackage{enumitem}
\usepackage{bm}             % Bold greek letters
\usepackage{mathtools} % For := sign
\usepackage{tikz}
\usetikzlibrary{shapes.geometric}
\usepackage{url}            % simple URL typesetting
\usepackage{caption}
\usepackage{subcaption}
\usepackage{pdfpages}
\usepackage{booktabs}
\ShortHeadings{CausalFairML via RPID}
{Bothmann, Dandl, \& Schomaker}
\firstpageno{1} % change upon acceptance

\begin{document}

% math spaces
\ifdefined\N                                                                
\renewcommand{\N}{\mathds{N}} % N, naturals
\else \newcommand{\N}{\mathds{N}} \fi 
\newcommand{\Z}{\mathds{Z}} % Z, integers
\newcommand{\Q}{\mathds{Q}} % Q, rationals
\newcommand{\R}{\mathds{R}} % R, reals
\ifdefined\C 
  \renewcommand{\C}{\mathds{C}} % C, complex
\else \newcommand{\C}{\mathds{C}} \fi
\newcommand{\continuous}{\mathcal{C}} % C, space of continuous functions
\newcommand{\M}{\mathcal{M}} % machine numbers
\newcommand{\epsm}{\epsilon_m} % maximum error

% counting / finite sets
\newcommand{\setzo}{\{0, 1\}} % set 0, 1
\newcommand{\setmp}{\{-1, +1\}} % set -1, 1
\newcommand{\unitint}{[0, 1]} % unit interval

% basic math stuff
\newcommand{\xt}{\tilde x} % x tilde
\newcommand{\argmax}{\operatorname{arg\,max}} % argmax
\newcommand{\argmin}{\operatorname{arg\,min}} % argmin
\newcommand{\argminlim}{\mathop{\mathrm{arg\,min}}\limits} % argmax with limits
\newcommand{\argmaxlim}{\mathop{\mathrm{arg\,max}}\limits} % argmin with limits  
\newcommand{\sign}{\operatorname{sign}} % sign, signum
\newcommand{\I}{\mathbb{I}} % I, indicator
\newcommand{\order}{\mathcal{O}} % O, order
\newcommand{\pd}[2]{\frac{\partial{#1}}{\partial #2}} % partial derivative
\newcommand{\floorlr}[1]{\left\lfloor #1 \right\rfloor} % floor
\newcommand{\ceillr}[1]{\left\lceil #1 \right\rceil} % ceiling

% sums and products
\newcommand{\sumin}{\sum\limits_{i=1}^n} % summation from i=1 to n
\newcommand{\sumim}{\sum\limits_{i=1}^m} % summation from i=1 to m
\newcommand{\sumjn}{\sum\limits_{j=1}^n} % summation from j=1 to p
\newcommand{\sumjp}{\sum\limits_{j=1}^p} % summation from j=1 to p
\newcommand{\sumik}{\sum\limits_{i=1}^k} % summation from i=1 to k
\newcommand{\sumkg}{\sum\limits_{k=1}^g} % summation from k=1 to g
\newcommand{\sumjg}{\sum\limits_{j=1}^g} % summation from j=1 to g
\newcommand{\meanin}{\frac{1}{n} \sum\limits_{i=1}^n} % mean from i=1 to n
\newcommand{\meanim}{\frac{1}{m} \sum\limits_{i=1}^m} % mean from i=1 to n
\newcommand{\meankg}{\frac{1}{g} \sum\limits_{k=1}^g} % mean from k=1 to g
\newcommand{\prodin}{\prod\limits_{i=1}^n} % product from i=1 to n
\newcommand{\prodkg}{\prod\limits_{k=1}^g} % product from k=1 to g
\newcommand{\prodjp}{\prod\limits_{j=1}^p} % product from j=1 to p

% linear algebra
\newcommand{\one}{\boldsymbol{1}} % 1, unitvector
\newcommand{\zero}{\mathbf{0}} % 0-vector
\newcommand{\id}{\boldsymbol{I}} % I, identity
\newcommand{\diag}{\operatorname{diag}} % diag, diagonal
\newcommand{\trace}{\operatorname{tr}} % tr, trace
\newcommand{\spn}{\operatorname{span}} % span
\newcommand{\scp}[2]{\left\langle #1, #2 \right\rangle} % <.,.>, scalarproduct
\newcommand{\mat}[1]{\begin{pmatrix} #1 \end{pmatrix}} % short pmatrix command
\newcommand{\Amat}{\mathbf{A}} % matrix A
\newcommand{\Deltab}{\mathbf{\Delta}} % error term for vectors

% basic probability + stats
\renewcommand{\P}{\mathds{P}} % P, probability
\newcommand{\E}{\mathds{E}} % E, expectation
\newcommand{\var}{\mathsf{Var}} % Var, variance
\newcommand{\cov}{\mathsf{Cov}} % Cov, covariance
\newcommand{\corr}{\mathsf{Corr}} % Corr, correlation
\newcommand{\normal}{\mathcal{N}} % N of the normal distribution
\newcommand{\iid}{\overset{i.i.d}{\sim}} % dist with i.i.d superscript
\newcommand{\distas}[1]{\overset{#1}{\sim}} % ... is distributed as ...

% machine learning
\newcommand{\Xspace}{\mathcal{X}} % X, input space
\newcommand{\Yspace}{\mathcal{Y}} % Y, output space
\newcommand{\nset}{\{1, \ldots, n\}} % set from 1 to n
\newcommand{\pset}{\{1, \ldots, p\}} % set from 1 to p
\newcommand{\gset}{\{1, \ldots, g\}} % set from 1 to g
\newcommand{\Pxy}{\mathbb{P}_{xy}} % P_xy
\newcommand{\Exy}{\mathbb{E}_{xy}} % E_xy: Expectation over random variables xy
\newcommand{\xv}{\mathbf{x}} % vector x (bold)
\newcommand{\xtil}{\tilde{\mathbf{x}}} % vector x-tilde (bold)
\newcommand{\yv}{\mathbf{y}} % vector y (bold)
\newcommand{\xy}{(\xv, y)} % observation (x, y)
\newcommand{\xvec}{\left(x_1, \ldots, x_p\right)^T} % (x1, ..., xp) 
\newcommand{\Xmat}{\mathbf{X}} % Design matrix
\newcommand{\allDatasets}{\mathds{D}} % The set of all datasets
\newcommand{\allDatasetsn}{\mathds{D}_n}  % The set of all datasets of size n 
\newcommand{\D}{\mathcal{D}} % D, data
\newcommand{\Dn}{\D_n} % D_n, data of size n
\newcommand{\Dtrain}{\mathcal{D}_{\text{train}}} % D_train, training set
\newcommand{\Dtest}{\mathcal{D}_{\text{test}}} % D_test, test set
\newcommand{\xyi}[1][i]{\left(\xv^{(#1)}, y^{(#1)}\right)} % (x^i, y^i), i-th observation
\newcommand{\Dset}{\left( \xyi[1], \ldots, \xyi[n]\right)} % {(x1,y1)), ..., (xn,yn)}, data
\newcommand{\defAllDatasetsn}{(\Xspace \times \Yspace)^n} % Def. of the set of all datasets of size n 
\newcommand{\defAllDatasets}{\bigcup_{n \in \N}(\Xspace \times \Yspace)^n} % Def. of the set of all datasets 
\newcommand{\xdat}{\left\{ \xv^{(1)}, \ldots, \xv^{(n)}\right\}} % {x1, ..., xn}, input data
\newcommand{\yvec}{\left(y^{(1)}, \hdots, y^{(n)}\right)^T} % (y1, ..., yn), vector of outcomes
\renewcommand{\xi}[1][i]{\xv^{(#1)}} % x^i, i-th observed value of x
\newcommand{\yi}[1][i]{y^{(#1)}} % y^i, i-th observed value of y 
\newcommand{\xivec}{\left(x^{(i)}_1, \ldots, x^{(i)}_p\right)^T} % (x1^i, ..., xp^i), i-th observation vector
\newcommand{\xj}{\xv_j} % x_j, j-th feature
\newcommand{\xjvec}{\left(x^{(1)}_j, \ldots, x^{(n)}_j\right)^T} % (x^1_j, ..., x^n_j), j-th feature vector
\newcommand{\phiv}{\mathbf{\phi}} % Basis transformation function phi
\newcommand{\phixi}{\mathbf{\phi}^{(i)}} % Basis transformation of xi: phi^i := phi(xi)

%%%%%% ml - models general
\newcommand{\lamv}{\bm{\lambda}} % lambda vector, hyperconfiguration vector
\newcommand{\Lam}{\bm{\Lambda}}	 % Lambda, space of all hpos
% Inducer / Inducing algorithm
\newcommand{\preimageInducer}{\left(\defAllDatasets\right)\times\Lam} % Set of all datasets times the hyperparameter space
\newcommand{\preimageInducerShort}{\allDatasets\times\Lam} % Set of all datasets times the hyperparameter space
% Inducer / Inducing algorithm
\newcommand{\ind}{\mathcal{I}} % Inducer, inducing algorithm, learning algorithm 

% continuous prediction function f
\newcommand{\ftrue}{f_{\text{true}}}  % True underlying function (if a statistical model is assumed)
\newcommand{\ftruex}{\ftrue(\xv)} % True underlying function (if a statistical model is assumed)
\newcommand{\fx}{f(\xv)} % f(x), continuous prediction function
\newcommand{\fdomains}{f: \Xspace \rightarrow \R^g} % f with domain and co-domain
\newcommand{\Hspace}{\mathcal{H}} % hypothesis space where f is from
\newcommand{\fbayes}{f^{\ast}} % Bayes-optimal model
\newcommand{\fxbayes}{f^{\ast}(\xv)} % Bayes-optimal model
\newcommand{\fkx}[1][k]{f_{#1}(\xv)} % f_j(x), discriminant component function
\newcommand{\fh}{\hat{f}} % f hat, estimated prediction function
\newcommand{\fxh}{\fh(\xv)} % fhat(x)
\newcommand{\fxt}{f(\xv ~|~ \thetab)} % f(x | theta)
\newcommand{\fxi}{f\left(\xv^{(i)}\right)} % f(x^(i))
\newcommand{\fxih}{\hat{f}\left(\xv^{(i)}\right)} % f(x^(i))
\newcommand{\fxit}{f\left(\xv^{(i)} ~|~ \thetab\right)} % f(x^(i) | theta)
\newcommand{\fhD}{\fh_{\D}} % fhat_D, estimate of f based on D
\newcommand{\fhDtrain}{\fh_{\Dtrain}} % fhat_Dtrain, estimate of f based on D
\newcommand{\fhDnlam}{\fh_{\Dn, \lamv}} %model learned on Dn with hp lambda
\newcommand{\fhDlam}{\fh_{\D, \lamv}} %model learned on D with hp lambda
\newcommand{\fhDnlams}{\fh_{\Dn, \lamv^\ast}} %model learned on Dn with optimal hp lambda 
\newcommand{\fhDlams}{\fh_{\D, \lamv^\ast}} %model learned on D with optimal hp lambda 

% discrete prediction function h
\newcommand{\hx}{h(\xv)} % h(x), discrete prediction function
\newcommand{\hh}{\hat{h}} % h hat
\newcommand{\hxh}{\hat{h}(\xv)} % hhat(x)
\newcommand{\hxt}{h(\xv | \thetab)} % h(x | theta)
\newcommand{\hxi}{h\left(\xi\right)} % h(x^(i))
\newcommand{\hxit}{h\left(\xi ~|~ \thetab\right)} % h(x^(i) | theta)
\newcommand{\hbayes}{h^{\ast}} % Bayes-optimal classification model
\newcommand{\hxbayes}{h^{\ast}(\xv)} % Bayes-optimal classification model

% yhat
\newcommand{\yh}{\hat{y}} % yhat for prediction of target
\newcommand{\yih}{\hat{y}^{(i)}} % yhat^(i) for prediction of ith targiet
\newcommand{\resi}{\yi- \yih}

% theta
\newcommand{\thetah}{\hat{\theta}} % theta hat
\newcommand{\thetab}{\bm{\theta}} % theta vector
\newcommand{\thetabh}{\bm{\hat\theta}} % theta vector hat
\newcommand{\thetat}[1][t]{\thetab^{[#1]}} % theta^[t] in optimization
\newcommand{\thetatn}[1][t]{\thetab^{[#1 +1]}} % theta^[t+1] in optimization
\newcommand{\thetahDnlam}{\thetabh_{\Dn, \lamv}} %theta learned on Dn with hp lambda
\newcommand{\thetahDlam}{\thetabh_{\D, \lamv}} %theta learned on D with hp lambda
\newcommand{\mint}{\min_{\thetab \in \Theta}} % min problem theta
\newcommand{\argmint}{\argmin_{\thetab \in \Theta}} % argmin theta

% densities + probabilities
% pdf of x 
\newcommand{\pdf}{p} % p
\newcommand{\pdfx}{p(\xv)} % p(x)
\newcommand{\pixt}{\pi(\xv~|~ \thetab)} % pi(x|theta), pdf of x given theta
\newcommand{\pixit}{\pi\left(\xi ~|~ \thetab\right)} % pi(x^i|theta), pdf of x given theta
\newcommand{\pixii}{\pi(\xi)} % pi(x^i), pdf of i-th x 
\newcommand{\pii}{\pi^{(i)}} % pi(x^i), pdf of i-th x 

% pdf of (x, y)
\newcommand{\pdfxy}{p(\xv,y)} % p(x, y)
\newcommand{\pdfxyt}{p(\xv, y ~|~ \thetab)} % p(x, y | theta)
\newcommand{\pdfxyit}{p\left(\xi, \yi ~|~ \thetab\right)} % p(x^(i), y^(i) | theta)

% pdf of x given y
\newcommand{\pdfxyk}[1][k]{p(\xv | y= #1)} % p(x | y = k)
\newcommand{\lpdfxyk}[1][k]{\log p(\xv | y= #1)} % log p(x | y = k)
\newcommand{\pdfxiyk}[1][k]{p\left(\xi | y= #1 \right)} % p(x^i | y = k)

% prior probabilities
\newcommand{\pik}[1][k]{\pi_{#1}} % pi_k, prior
\newcommand{\lpik}[1][k]{\log \pi_{#1}} % log pi_k, log of the prior
\newcommand{\pit}{\pi(\thetab)} % Prior probability of parameter theta

% posterior probabilities
\newcommand{\post}{\P(y = 1 ~|~ \xv)} % P(y = 1 | x), post. prob for y=1
\newcommand{\postk}[1][k]{\P(y = #1 ~|~ \xv)} % P(y = k | y), post. prob for y=k
\newcommand{\pidomains}{\pi: \Xspace \rightarrow \unitint} % pi with domain and co-domain
\newcommand{\pibayes}{\pi^{\ast}} % Bayes-optimal classification model
\newcommand{\pixbayes}{\pi^{\ast}(\xv)} % Bayes-optimal classification model
\newcommand{\pix}{\pi(\xv)} % pi(x), P(y = 1 | x)
\newcommand{\pikx}[1][k]{\pi_{#1}(\xv)} % pi_k(x), P(y = k | x)
\newcommand{\pikxt}[1][k]{\pi_{#1}(\xv ~|~ \thetab)} % pi_k(x | theta), P(y = k | x, theta)
\newcommand{\pixh}{\hat \pi(\xv)} % pi(x) hat, P(y = 1 | x) hat
\newcommand{\pikxh}[1][k]{\hat \pi_{#1}(\xv)} % pi_k(x) hat, P(y = k | x) hat
\newcommand{\pixih}{\hat \pi(\xi)} % pi(x^(i)) with hat
\newcommand{\pikxih}[1][k]{\hat \pi_{#1}(\xi)} % pi_k(x^(i)) with hat
\newcommand{\pdfygxt}{p(y ~|~\xv, \thetab)} % p(y | x, theta)
\newcommand{\pdfyigxit}{p\left(\yi ~|~\xi, \thetab\right)} % p(y^i |x^i, theta)
\newcommand{\lpdfygxt}{\log \pdfygxt } % log p(y | x, theta)
\newcommand{\lpdfyigxit}{\log \pdfyigxit} % log p(y^i |x^i, theta)

% probababilistic
\newcommand{\bayesrulek}[1][k]{\frac{\P(\xv | y= #1) \P(y= #1)}{\P(\xv)}} % Bayes rule
\newcommand{\muk}{\bm{\mu_k}} % mean vector of class-k Gaussian (discr analysis) 

% residual and margin
\newcommand{\eps}{\epsilon} % residual, stochastic
\newcommand{\epsi}{\epsilon^{(i)}} % epsilon^i, residual, stochastic
\newcommand{\epsh}{\hat{\epsilon}} % residual, estimated
\newcommand{\yf}{y \fx} % y f(x), margin
\newcommand{\yfi}{\yi \fxi} % y^i f(x^i), margin
\newcommand{\Sigmah}{\hat \Sigma} % estimated covariance matrix
\newcommand{\Sigmahj}{\hat \Sigma_j} % estimated covariance matrix for the j-th class

% ml - loss, risk, likelihood
\newcommand{\Lyf}{L\left(y, f\right)} % L(y, f), loss function
\newcommand{\Lxy}{L\left(y, \fx\right)} % L(y, f(x)), loss function
\newcommand{\Lxyi}{L\left(\yi, \fxi\right)} % loss of observation
\newcommand{\Lxyt}{L\left(y, \fxt\right)} % loss with f parameterized
\newcommand{\Lxyit}{L\left(\yi, \fxit\right)} % loss of observation with f parameterized
\newcommand{\Lxym}{L\left(\yi, f\left(\bm{\tilde{x}}^{(i)} ~|~ \thetab\right)\right)} % loss of observation with f parameterized
\newcommand{\Lpixy}{L\left(y, \pix\right)} % loss in classification
\newcommand{\Lpixyi}{L\left(\yi, \pixii\right)} % loss of observation in classification
\newcommand{\Lpixyt}{L\left(y, \pixt\right)} % loss with pi parameterized
\newcommand{\Lpixyit}{L\left(\yi, \pixit\right)} % loss of observation with pi parameterized
\newcommand{\Lhxy}{L\left(y, \hx\right)} % L(y, h(x)), loss function on discrete classes
\newcommand{\Lr}{L\left(r\right)} % L(r), loss defined on residual (reg) / margin (classif)
\newcommand{\lone}{|y - \fx|} % L1 loss
\newcommand{\ltwo}{\left(y - \fx\right)^2} % L2 loss
\newcommand{\lbernoullimp}{\ln(1 + \exp(-y \cdot \fx))} % Bernoulli loss for -1, +1 encoding
\newcommand{\lbernoullizo}{- y \cdot \fx + \log(1 + \exp(\fx))} % Bernoulli loss for 0, 1 encoding
\newcommand{\lcrossent}{- y \log \left(\pix\right) - (1 - y) \log \left(1 - \pix\right)} % cross-entropy loss
\newcommand{\lbrier}{\left(\pix - y \right)^2} % Brier score
\newcommand{\risk}{\mathcal{R}} % R, risk
\newcommand{\riskbayes}{\mathcal{R}^\ast}
\newcommand{\riskf}{\risk(f)} % R(f), risk
\newcommand{\riskdef}{\E_{y|\xv}\left(\Lxy \right)} % risk def (expected loss)
\newcommand{\riskt}{\mathcal{R}(\thetab)} % R(theta), risk
\newcommand{\riske}{\mathcal{R}_{\text{emp}}} % R_emp, empirical risk w/o factor 1 / n
\newcommand{\riskeb}{\bar{\mathcal{R}}_{\text{emp}}} % R_emp, empirical risk w/ factor 1 / n
\newcommand{\riskef}{\riske(f)} % R_emp(f)
\newcommand{\risket}{\mathcal{R}_{\text{emp}}(\thetab)} % R_emp(theta)
\newcommand{\riskr}{\mathcal{R}_{\text{reg}}} % R_reg, regularized risk
\newcommand{\riskrt}{\mathcal{R}_{\text{reg}}(\thetab)} % R_reg(theta)
\newcommand{\riskrf}{\riskr(f)} % R_reg(f)
\newcommand{\riskrth}{\hat{\mathcal{R}}_{\text{reg}}(\thetab)} % hat R_reg(theta)
\newcommand{\risketh}{\hat{\mathcal{R}}_{\text{emp}}(\thetab)} % hat R_emp(theta)
\newcommand{\LL}{\mathcal{L}} % L, likelihood
\newcommand{\LLt}{\mathcal{L}(\thetab)} % L(theta), likelihood
\newcommand{\LLtx}{\mathcal{L}(\thetab | \xv)} % L(theta|x), likelihood
\newcommand{\logl}{\ell} % l, log-likelihood
\newcommand{\loglt}{\logl(\thetab)} % l(theta), log-likelihood
\newcommand{\logltx}{\logl(\thetab | \xv)} % l(theta|x), log-likelihood
\newcommand{\errtrain}{\text{err}_{\text{train}}} % training error
\newcommand{\errtest}{\text{err}_{\text{test}}} % test error
\newcommand{\errexp}{\overline{\text{err}_{\text{test}}}} % avg training error

% lm
\newcommand{\thx}{\thetab^T \xv} % linear model
\newcommand{\olsest}{(\Xmat^T \Xmat)^{-1} \Xmat^T \yv} % OLS estimator in LM 

\newcommand{\xti}{\xtil^{(i)}}
\newcommand{\yti}{\tilde{y}^{(i)}}
\newcommand{\yt}{\tilde{y}}
\newcommand{\xtilde}{\tilde{\xv}}

\newcommand{\Xtil}{\tilde{X}}
\newcommand{\Xtj}{\Xtil_j}

\newcommand{\pitil}{\phi}
\newcommand{\piti}{\pitil^{(i)}}
\newcommand{\pitj}{\pitil^{(j)}}
\newcommand{\pitxtih}{\hat{\pitil}(\xti)} 
\newcommand{\pitxih}{\hat{\pitil}(\xi)} 
\newcommand{\pitxti}{\pitil(\xti)}
\newcommand{\pitxi}{\pitil(\xi)}
\newcommand{\pixv}{\pi(\xv)}

\newcommand{\spiti}{s(\piti)}
\newcommand{\spitxtih}{s(\pitxtih)}
\newcommand{\spitxih}{s(\pitxih)}
\newcommand{\spitxti}{s(\pitxti)}
\newcommand{\spitxi}{s(\pitxi)}

\newcommand{\spii}{s(\pii)}
\newcommand{\spixih}{s(\pixih)}
\newcommand{\spixii}{s(\pixii)}

\newcommand{\ti}{t^{(i)}}
\newcommand{\tj}{t^{(j)}}
\newcommand{\wi}{m^{(i)}}
\newcommand{\wj}{m^{(j)}}
\newcommand{\ui}{z^{(i)}}
\newcommand{\vi}{v^{(i)}}
\newcommand{\mui}{\mu^{(i)}}

\newcommand{\sfu}{s(\cdot)}
\newcommand{\pij}{\pi^{(j)}}
\newcommand{\pif}{\pi(\cdot)}
\newcommand{\pih}{\hat{\pi}}
\newcommand{\pihi}{\pih^{(i)}}
\newcommand{\pihj}{\pih^{(j)}}
\newcommand{\pihf}{\pih(\cdot)}
\newcommand{\pitf}{\pitil(\cdot)}
\newcommand{\pith}{\hat{\pitil}}
\newcommand{\pithf}{\hat{\pitil}(\cdot)}
\newcommand{\pithi}{\hat{\pitil}^{(i)}}
\newcommand{\pithj}{\hat{\pitil}^{(j)}}

\newcommand{\psih}{\hat{\psi}}
\newcommand{\psihi}{\psih^{(i)}}

\newcommand{\pim}{\pi^{m}}
\newcommand{\pifem}{\pi^{f}}

\newcommand{\thetav}{\bm{\theta}}
\newcommand{\betav}{\bm{\beta}}
\newcommand{\betavh}{\bm{\hat{\beta}}}
\newcommand{\zv}{\mathbf{z}}
\newcommand{\zi}[1][i]{\zv^{(#1)}} % z^i, i-th observed value of z
\newcommand{\zvh}{\hat{\zv}}
\newcommand{\zhi}[1][i]{\zvh^{(#1)}}
\newcommand{\betah}{\hat{\beta}}

\newcommand{\pa}{\textnormal{PA}}
\newcommand{\pax}{\textnormal{PA}_{\mathcal{X}}}
\newcommand{\paz}{\textnormal{PA}_{\mathcal{Z}}}

\newcommand{\xyti}[1][i]{\left(\xtil^{(#1)}, \yt^{(#1)}\right)} % (x^i, y^i), i-th observation
\newcommand{\Dsett}{\left( \xyti[1], \ldots, \xyti[n]\right)} % {(x1,y1)), ..., (xn,yn)}, data
\newcommand{\xyhi}[1][i]{\left(\hat{\xv}^{(#1)}, \hat{y}^{(#1)}\right)} % (x^i, y^i), i-th observation
\newcommand{\Dsetw}{\left( \xyhi[1], \ldots, \xyhi[n]\right)} % {(x1,y1)), ..., (xn,yn)}, data
\newcommand{\Dt}{\tilde{\mathcal{D}}} % D, data

\newcommand{\Xspacet}{\tilde{\Xspace}}

% Farbige Kommentare
\newcommand{\lb}[1]{\textcolor{blue}{[L: #1]}}
\newcommand{\kp}[1]{\textcolor{blue}{[K: #1]}}
\newcommand{\ms}[1]{\textcolor{violet}{[M: #1]}}
\newcommand{\sd}[1]{\textcolor{orange}{[S: #1]}}
\newcommand{\bp}[1]{\textcolor{blue}{$\rightarrow$ #1 \\}}

% Bold oder italic
\newcommand{\myemph}[1]{\textit{#1}}

% Intervention
\newcommand{\txai}{\tilde{x}_{A}^{(i)}}
\newcommand{\txsi}{\tilde{x}_{S}^{(i)}}
\newcommand{\txri}{\tilde{x}_{R}^{(i)}}
\newcommand{\txii}{\tilde{x}_{I}^{(i)}}

\newcommand{\tyi}{\tilde{y}^{(i)}}

\title{Causal Fair Machine Learning via Rank-Preserving Interventional Distributions}

%\author{\name Anonymous Authors}
\author{\name Ludwig Bothmann \email ludwig.bothmann@lmu.de \\
       \addr LMU Munich, Germany\\
       Munich Center for Machine Learning (MCML), Germany
       \AND
       \name Susanne Dandl \email susanne.dandl@uzh.ch \\
       \addr University of Zurich, Switzerland
       \AND
       \name Michael Schomaker \email michael.schomaker@stat.uni-muenchen.de \\
       \addr LMU Munich, Germany\\
       University of Cape Town, South Africa
       }

% For research notes, remove the comment character in the line below.
% \researchnote

\maketitle

\begin{abstract}
  A decision can be defined as fair if equal individuals are treated equally and unequals unequally. 
  Adopting this definition, the task of designing machine learning (ML) models that mitigate unfairness in automated decision-making systems must include causal thinking when introducing protected attributes: Following a recent proposal, we define individuals as being normatively equal if they are equal in a fictitious, normatively desired (FiND) world, where the protected attributes have no (direct or indirect) causal effect on the target. 
  We propose \myemph{rank-preserving interventional distributions} to define a specific FiND world in which this holds and a \myemph{warping method} for estimation. Evaluation criteria for both the method and the resulting ML model are presented and validated through simulations. 
  Experiments on empirical data showcase the practical application of our method and compare results with ``fairadapt'' (Plečko and Meinshausen, 2020), a different approach for mitigating unfairness by causally preprocessing data that uses quantile regression forests. With this, we show that our warping approach effectively identifies the most discriminated individuals and mitigates unfairness. %\footnote{Transparency note: This manuscript is an extended version of a paper presented at a workshop in autumn 2023. Major differences to the workshop paper: Extended explanation of the warping approach, including a visualization to help intuition; including analysis of German Credit data in the experiments section; comparison with \textit{fairadapt}: theoretical contrasting and experiments on German Credit data; additionally comparing fairness through unawareness in the appendix; ethical statements.
%  To preserve anonymity, bibliographic details will be added upon acceptance.}
\end{abstract}

%%%%%%%%%%%%% CONTENT
\section{Introduction}
\label{sec:intro}

Automated decision-making (ADM) systems can support human decision-makers by predicting some variable of interest via a machine learning (ML) model. The data used for learning such ML models can have historical bias, i.e., show normatively undesirable discrimination against certain groups of protected attributes (PAs). When left unaddressed, this historical bias leads to biased ML models, generating fairness problems in ADM systems.
The research field of fair machine learning (fairML) has quickly grown around this problem in recent years, giving birth to various ``fairness metrics'' \shortcite<such as, e.g., demographic parity, see>[for an overview]{barocas_fairness_2019}.
%However, as \citep{anonymous_2024} showed recently, these fairness metrics often lack a solid philosophical foundation making it unclear which concept of fairness is measured by the respective metrics.

A critique raised by \shortciteA{bothmann_what_2024} is that the question of \myemph{what fairness is} -- as a philosophical concept -- is rarely discussed. 
%\footnote{To preserve anonymity, we added a blinded version of this paper in the appendix. Upon acceptance, we remove this and update the citation.}
Hence, the proposed fairness metrics lack a philosophical justification, making it unclear which concept of fairness is measured by the respective metrics.
The authors propose a consistent concept of fairness and outline how this should be integrated into the design of ML models in ADM systems.
Following \citeA{aristotle_nicomachean_2009}, they define a treatment as being fair ``if equals are treated equally and if unequals are treated unequally'', where equality is a task-specific notion. More concretely, they adopt the view of ``task-specific merit'' which captures Aristotle's concept of ``worthiness''. In their framework (expanding on Aristotle's idea of geometric proportionality), a treatment $\ti$ of an individual $i$ is considered fair, iff it follows a normative treatment function $s(\cdot)$ of this task-specific merit $\wi$, i.e., iff $\ti = s(\wi)$. Suppose the task-specific merit $\wi$ involves quantities that cannot be measured at the decision time (such as the probability of paying back a credit). In that case, an attempt to estimate these quantities by an ML model can be made, imposing quality requirements upon the ML model.
Specifically, they distinguish between descriptively unfair treatment and normatively unfair treatment. The former can already happen without PAs, if the ML model is not ``individually well-calibrated'', essentially meaning perfect predictions -- which can be relaxed to a sensible tolerable error. The latter is a causal notion where the PAs change the definition of the task-specific merit $\wi$ normatively: They conceive a \myemph{fictitious, normatively desired (FiND) world}, where the PA has no (direct or indirect) causal effect on the target variable. Individuals are then normatively considered equal if they are equal with respect to the task-specific merit in the FiND world. %, and leave concrete algorithms to future research.

We build on this work by proposing concrete intervention mechanisms that remove direct and indirect effects so that one can target appropriate estimands and derive estimation procedures.
%as well as by proposing evaluation methods and by presenting experimental and real-world results.
As a starting point, we define a directed acyclic graph (DAG) that describes the causal relations in the real world. %The DAG might be designed by pure expert knowledge or supported by methods of causal discovery \citep[see, e.g.,][for an overview]{nogueira_methods_2022}.
The DAG in the FiND world is then created by deleting all arrows that constitute paths from the PA to the target. %, see Figure \ref{fig:DAG} for an example.
We achieve this through specific interventions on the descendants of the PA on those paths, leading to \myemph{rank-preserving interventional distributions} (RPID).
%on the mediators, such that the mediator distributions do not differ with respect to the groups defined by the PA.
This intervention is rank-preserving in the following sense: Individuals of the disadvantaged group maintain the rank they have in the real world (compared with other individuals of the disadvantaged group) as population-wide rank in the FiND world (compared with all individuals), see Section \ref{sec:estimand}.
%This intervention is rank-preserving in the sense that individuals of the disadvantaged group maintain the group-specific real-world rank (in variables that are intervened on) as population-wide FiND-world rank (see Section \ref{sec:estimand}).
%While there is a growing body of literature for path-specific effects, both in philosophy \citep[e.g.,][]{weinberger_path-specific_2019} and ML \citep[e.g.,][]{chikahara_learning_2021}, we believe that this approach of deleting all the causal effects of the PA resembles more closely the legal necessities of not discriminating based on the PAs. %(however, we would be very interested in a discussion on this question at the workshop).

After identifying the estimand, we propose a \myemph{warping method} for estimation that maps real-world data to a \myemph{warped world} which in turn approximates the FiND world, see Section \ref{sec:training}.
We call this a \myemph{quasi-individual} approach because individual ``merits'' are pulled through to the warped world. % via the quantiles of the residuals.
Finally, an ML model is trained on the warped data that can be used at prediction time after warping a new observation.
Since final prediction models are trained and evaluated in the warped world, our approach can be considered to be a \myemph{pre-processing approach} \cite<see>[for a categorization of different approaches in fairML]{caton_fairness_2023}.
%For a new observation at prediction time, first, the warping (as learned during training) is applied, and then the target is predicted with the warped data ML model.

We propose evaluation metrics both for evaluating the warping method in a simulation study and for evaluating an ML model using warped data in an applied use case in Section \ref{sec:evaluation}. With these evaluation methods, it is possible to (i) describe the degree to which a given individual is discriminated against in the real world, (ii) identify the individuals that profit or suffer most from PA-related discrimination in the real world and (iii) identify features that are (globally) most relevant for the discrimination in the real world.
In a simulation study, we show that our warping method can approximate the FiND world, identify the most discriminated individuals, and eliminate the effects of the PA in the warped world (Sections \ref{sec:sim} and \ref{sec:sim-results}).
Finally, we apply the proposed methodology to German Credit data \cite{hofmann_statlog_1994}, showing how to use our framework in practice to mitigate PA-related discrimination and to identify the most strongly discriminated individuals (Section \ref{sec:results_german}). In all experiments, we compare RPID with \textit{fairadapt} \cite{plecko_fair_2020}, which can be considered as an alternative warping approach, see descriptions in Sections \ref{sec:related-work} and \ref{sec:experiments}.

Our method has two main applications: (i) viewed as a pre-processing approach, it can be used to account ``for historical inequalities which actively ought to be eroded'' \shortcite{wachter_bias_2021} -- by warping real-world data towards a FiND world, training ML models with the pre-processed data and predict from these models; (ii) viewed as an evaluation method, it can be used to search for individuals that are discriminated against most strongly in the real world -- by comparing ML model predictions in the real world and the warped world.

\section{Related Work}
\label{sec:related-work}

In addition to \myemph{group fairness} concepts \cite<see, e.g.,>[for an overview]{verma_fairness_2018}, approaches of (non-causal) \myemph{individual fairness} have been proposed, starting with \shortciteA{dwork_fairness_2012}, who require that similar individuals should be treated similarly \cite<see also>{bechavod_metric-free_2020,chouldechova_frontiers_2018,friedler_impossibility_2016}.
An early notion of \myemph{causal fairness} was made by \shortciteA{kusner_counterfactual_2017}, who conceive a fictitious world where an individual belongs to a different subgroup of the PA, defining a decision as fair if it is equal in the real and fictitious world. As \citeA[Section 3.4]{bothmann_what_2024} elaborate on, this idea differs -- while seemingly similar at first glance -- substantially from conceiving a FiND world and approximating it with a warped world, since the (fictitious) decisions in the real and warped world do not have to be the same.
Including causality in the fairness debate and conceiving a fictitious world was also proposed by, e.g., \shortciteA{zhang_equality_2018,zhang_fairness_2018,nabi_fair_2018,nabi_learning_2019,nabi_optimal_2022,pfohl_counterfactual_2019}, where different ideas underlie those fictitious worlds.

% Path-Specific
With the fairness concept introduced by \citeA{bothmann_what_2024}, real world and FiND world are distinguished by the idea that in the FiND world, there must be no causal effects from the PA on the target -- neither indirectly, nor directly. This means that all arrows starting in the PA and eventually leading to the target variable are deleted by a specific, meaningful intervention that takes into account the concept of equal treatment of normatively equal individuals (dashed arrows in Figure \ref{fig:DAG}). This idea differs from what the literature on path-specific effects \shortcite{chiappa_path-specific_2019,chikahara_learning_2021,weinberger_path-specific_2019} conceives. However, we believe that this more adequately captures the legal requirements of many laws that demand that individuals must not be discriminated against based on the PA\footnote{E.g., %US Civil Rights Act of 1964: \url{https://www.dol.gov/agencies/oasam/civil-rights-center/statutes/civil-rights-act-of-1964} or 
Charter of Fundamental Rights of the European Union: \url{https://www.citizensinformation.ie/en/government-in-ireland/european-government/eu-law/charter-of-fundamental-rights/}} -- rendering it irrelevant which path in the DAG this effect follows.

% Pre-Processing
Warping real-world data with RPID and subsequently training an ML model on the warped data can be seen as a pre-processing approach. While the number of proposed pre-processing approaches is steadily growing \shortcite<see>[for an overview]{caton_fairness_2023}, to the best of our knowledge, there is just one approach that is close enough to RPID to justify concrete comparisons in this article: \shortciteA{plecko_fair_2020} propose to compute \textit{fair twins} for each observation changing the PA to a baseline value using a \textit{do}-intervention on the PA. In Section \ref{sec:experiments} we elaborate further on their idea and compare it in experiments with RPID.

% Philosophy
From a legal and philosophical perspective, \shortciteA{wachter_bias_2021} discuss the two notions \myemph{formal equality} and \myemph{substantive equality} of EU non-discrimination law. They relate this to the concepts of \myemph{bias preserving fairness metrics} and \myemph{bias transforming fairness metrics}, respectively. With the concept of conceiving a FiND world, we are in the realm of substantive equality and bias transforming fairness metrics, since the normatively unwanted discriminations in the real world are actively tackled and mitigated via warping.
A similar perspective is adopted by \shortciteA{alvarez_counterfactual_2023} who propose to evaluate ML models by \myemph{counterfactual situation testing}. 
Their basic idea is to not compare the prediction of an individual of a PA-group, e.g., females, with predictions of members of the other PA-group, e.g., males, that have similar features, but to first compute a counterfactual for that female individual and to search for male comparisons in the neighborhood of that counterfactual. While this reflects a comparable concept of a FiND world where the causal effects of PA are removed, their approach differs from the approach presented here since no models are trained with the counterfactuals, but they are just used for evaluation.

% The modification of causal effects through deleting arrows by particular interventions has been criticized by \cite{hu_whats_2020}, claiming that these effects are ``constitutive features'' of the PA that cannot be removed without stripping the PA from the ``meaning [of this variable] in our world''.
% We argue that this graph surgery is a valid method to conceive a FiND world, where the PA has no causal effect on the target. In the FiND world, the PA does not have to have the same meaning as in our real world -- it is more about the idea that people are normatively considered equal if they only differ in the PA, and hence are not distinguishable in the FiND world.

\section{Methods}
\label{sec:methods}

As derived in \citeA{bothmann_what_2024}, in order to derive the decision basis for fair decisions, we must conceive a ``fictitious, normatively desired (FiND) world in which the PA has no causal effect'' on the target variable, ``neither directly nor indirectly''. In the following, we adopt this idea, elaborate it further by concretely specifying causal and statistical estimands, and derive an estimation method, thereby building concrete and actionable algorithms for approximating the FiND world by a ``warped world'' and for using Causal Fair ML (cfML) in applied use-cases.\footnote{For a thorough explanation of the philosophical foundations of FiND and warped world, see again \shortciteA{bothmann_what_2024}.}

Our method consists of four fundamental steps: (i) We first define the estimand as the joint distribution in the FiND world, described by specific interventions, tied to so-called \myemph{rank-preserving interventional distributions} (RPID, see Section \ref{sec:estimand});
%(ii) estimation by learning ``warping models'' to translate real-world data into the warped world and by training an ML model (for predicting the target) in the warped world,
(ii) We then estimate the joint/conditional distributions of interest in the FiND world, based on a specific $g$-formula type of factorization that follows from the respective structural assumptions encoded in a directed acyclic graph and allows us to ``warp'' the real-world data into the warped world (see Section \ref{sec:warping}); With this, we can (iii) train an ML model (for predicting the target) in this warped world
(see Section \ref{sec:training_warped}), and (iv) predict on a new observation in the warped world using the above warping models and ML model (see Section \ref{sec:prediction}).

\subsection{Estimand}
\label{sec:estimand}

Before we describe estimation in Section \ref{sec:training} and prediction in Section \ref{sec:prediction}, we start by defining the estimands.

\subsubsection{DAGs in the Real and the FiND World}
\label{sec:dag}

Deriving a DAG falls into the realm of \myemph{causal discovery} \shortcite<see, e.g.,>[for a review of current methods]{nogueira_methods_2022}. Since this is a notoriously hard challenge in practice, the alternative is to define the DAG with expert knowledge, as is typically done in epidemiology and medicine, where knowledge from human decision-makers is readily available \cite{hernan_causal_2020}. In the remainder, we are agnostic to the question of how the DAG was constructed and will assume that all DAGs are correct;
%in the sense that they mirror the causal relationships in the real world between the features shown in the DAG;
note that this may be an optimistic (and untestable) assumption and can hamper success in practice.

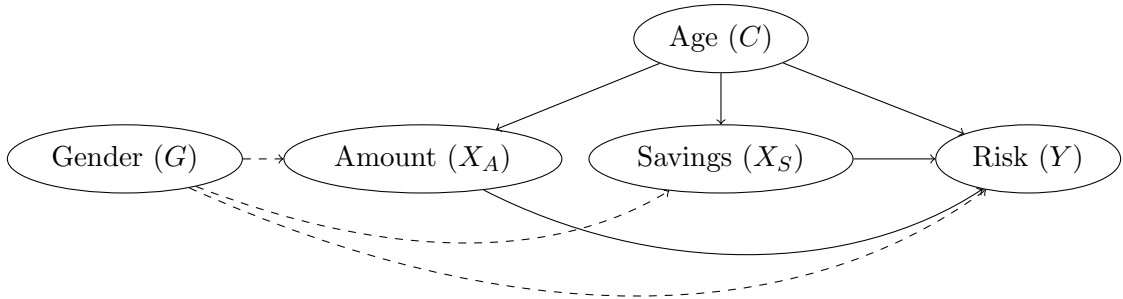
\begin{figure}[ht]
    \centering
\begin{tikzpicture}[x=6in, y=5in]
\node[ellipse, draw] (v0) at (0.600,-0.417) {Age ($C$)};
\node[ellipse, draw] (v1) at (0.340,-0.543) {Amount ($X_A$)};
\node[ellipse, draw] (v2) at (0.08,-0.543) {Gender ($G$)};
\node[ellipse, draw] (v3) at (0.600,-0.543) {Savings ($X_S$)};
\node[ellipse, draw] (v4) at (0.868,-0.543) {Risk ($Y$)};
\draw [->] (v0) edge (v1);
\draw [->] (v0) edge (v3);
\draw [->] (v0) edge (v4);
%\draw [->] (v1) edge (v4);
\draw [->] (v1) .. controls (0.540,-0.666) and (0.713,-0.666) .. (v4);
\draw [dashed, ->] (v2) edge (v1);
\draw [dashed, ->] (v2) .. controls (0.310,-0.65) and (0.455,-0.65) .. (v3);
\draw [dashed, ->] (v2) .. controls (0.415,-0.724) and (0.662,-0.724) .. (v4);
\draw [->] (v3) edge (v4);
\end{tikzpicture}
    \caption{Assumed DAGs for credit risk example. In the FiND world, only solid arrows exist, and in the real world, all arrows exist.}
    \label{fig:DAG}
\end{figure}

Two DAGs must be defined: the DAG in the real world and the DAG in the FiND world -- where the PAs have no causal effect on the target. Figure \ref{fig:DAG} shows the two DAGs that we assume for the example of the German Credit data \cite{hofmann_statlog_1994}. Note that these DAGs are chosen for illustrative purposes and not because there is empirical evidence or expert knowledge that justifies exactly those DAGs. We reduced the feature set for a clearer presentation: Age (a confounder $C$) is the numerical age of an individual; Gender (the PA $G$) is assumed to be binary (classes \textit{female} and \textit{male}) in the remainder for the ease of presentation, but note that an extension on multi-categorical gender is methodologically straightforward;
%(in practice, we could run into problems of high estimation errors for small classes, though);
Amount (feature $X_A$) is the amount of credit applied for; 
Savings (feature $X_S$) is a binary variable, indicating if the person has small savings (1) or not (0); 
and Risk (target $Y$) is the binary risk category with values \textit{good} (1) and \textit{bad} (0).

\subsubsection{Rank-Preserving Interventional Distributions}

Several possible interventions can delete the dashed arrows in Figure \ref{fig:DAG} and thus lead to the FiND world DAG. We propose the following idea of ``rank-preserving interventional distributions'' (RPID), which we believe is an appropriate way to define these interventions when the goal is to mitigate unfairness, for the following reasons:
\begin{enumerate}
\item \myemph{Fair world assurance}: They guarantee a world, in which the PA has no direct or indirect effect on the outcome.
\item \myemph{Philosophical soundness}: They do not require a conceptualization of an intervention on the PA \shortcite<an intervention on PAs such as gender or race has faced significant philosophical critique, see, e.g.,>{issa_kohler-hausmann_eddie_2019}.
\item \myemph{Quasi-individual fairness}: They aim for quasi-individual fairness, recognizing that exact individual fairness is unattainable because individual causal effects (i.e., individual differences between FiND world outcomes and observed outcomes) can never be identified \shortcite{hernan_causal_2020}. By targeting the quantiles of a specific distribution, we achieve ``almost-individual'' fairness in a rank-preserving sense, as explained below. 
\end{enumerate}
To start with, we assume that the given DAGs (as shown in Figure \ref{fig:DAG}) correctly mirror the causal relationships in both the real world and the FiND world. 
%(with full acknowledgment that this assumption might be wrong) 
%and use the DAGs shown in Figure \ref{fig:DAG} as an illustrative example. 
Slightly adapting the notation and terminology of \citeA{pearl_causality_2009}\footnote{Definition 7.1.1, p.\ 203}, a general structural causal model (SCM) is given by
% XXX
\begin{equation*}
    X_j \coloneqq f_j(pa(X_j), U_j), \quad j \in \pset,
\end{equation*}

\noindent where $U_1, \dots, U_p$ denote exogeneous independent random variables, and $pa(X)$ are parent nodes of $X$. In our example, the SCM in the real world (i.e., pre-intervention) is given by %equations
\begin{align*}
    G & \coloneqq f(U_G) \\
    C & \coloneqq f(U_C) \\
    X_A &\coloneqq f_A(G, C, U_A) \\
    X_S &\coloneqq f_S(G, C, U_S) \\
    Y &\coloneqq f_Y(G, C, X_A, X_S, U_Y),
\end{align*}

\noindent which entails a joint distribution that can be factorized according to our working order:
\begin{equation}\label{formula:pre_int_fact}
P(Y, X_S, X_A, C, G) = P_Y(Y|X_S, X_A, C, G) P_S(X_S|C, G) P_A(X_A|C, G) P_C(C) P_G(G).
\end{equation}
%Generated by this causal model, we assume to have a dataset $\D = \Dset$. For finding a mapping from real world to warped world, i.e., $m_y: \Yspace \rightarrow \Yspace, \ y \mapsto \yt$ and $m_\xv: \Xspace \rightarrow \Xspacet, \ \xv \mapsto \xtilde$, we have to make all descendants from the PA neutral w.r.t.\ the PA. That is, we correct the values of $\xv$ and $y$ by taking out the effect of the PA.
For the FiND world, we must make all descendants from the PA neutral w.r.t.~the PA.
We achieve this by a fictitious intervention rule $d_p$ on the mediators and outcome only, i.e., no ``modification'' of the potentially sensitive PA is required (Eq.~\ref{eq:intervention}).

\begin{flalign}
d_p %= d(G,C,\tilde{q}) 
&=\left\{ \begin{array}{cl}
               X_{A}^{(i)}=\txai  & \text{where} ~ \txai~ \text{is the (} p_A^{(i)}\times100) \text{\% quantile of the conditional} \\& \text{mediator distribution among the reference PA value, i.e.,} \\& P(X_A \leq \txai|C=c^{(i)},G=m) = p_A^{(i)},~ \text{and} ~p_A^{(i)}~ \text{is determined}\\  &\text{by the pre-intervention quantile of unit $i$, i.e.,} \\& p_A^{(i)} = P(X_A \leq x_{A}^{(i)} \mid C=c^{(i)},G=g^{(i)}).\\
               &\\
               X_{S}^{(i)}=\txsi  & \text{where} ~ \txsi~ \text{is the (} p_S^{(i)}\times100) \text{\% quantile of the conditional} \\& \text{mediator distribution among the reference PA value, i.e.,} \\& P(X_S \leq \txsi | C=c^{(i)},G=m) = p_S^{(i)},~ \text{and} ~p_S^{(i)}~ \text{is determined}\\  &\text{by the pre-intervention quantile of unit $i$, i.e.,} \\& p_S^{(i)} = P(X_S \leq x_{S}^{(i)}| C=c^{(i)},G=g^{(i)}).\\
               &\\ 
               Y^{(i)}=\tyi  & \text{where} ~ \tyi~ \text{is the (} p_Y^{(i)}\times100) \text{\% quantile of the counter\-fac\-tual} \\& \text{outcome distribution for the reference PA value, i.e.,} \\& P(Y \leq \tyi \mid \txai,\txsi,C=c^{(i)},G=m) = p_Y^{(i)},~ \text{and} ~p_Y^{(i)}~ \text{is}\\  &\text{based on the pre-intervention quantile of unit $i$, i.e.,} \\& p_Y^{(i)} = P(Y \leq y^{(i)} \mid X_A=x_{A}^{(i)},X_S=x_{S}^{(i)},C=c^{(i)},G=g^{(i)}).\\
               \end{array} 
               \right. & \label{eq:intervention}
\end{flalign}

\begin{figure}
    \centering
    \includegraphics[width=\textwidth]{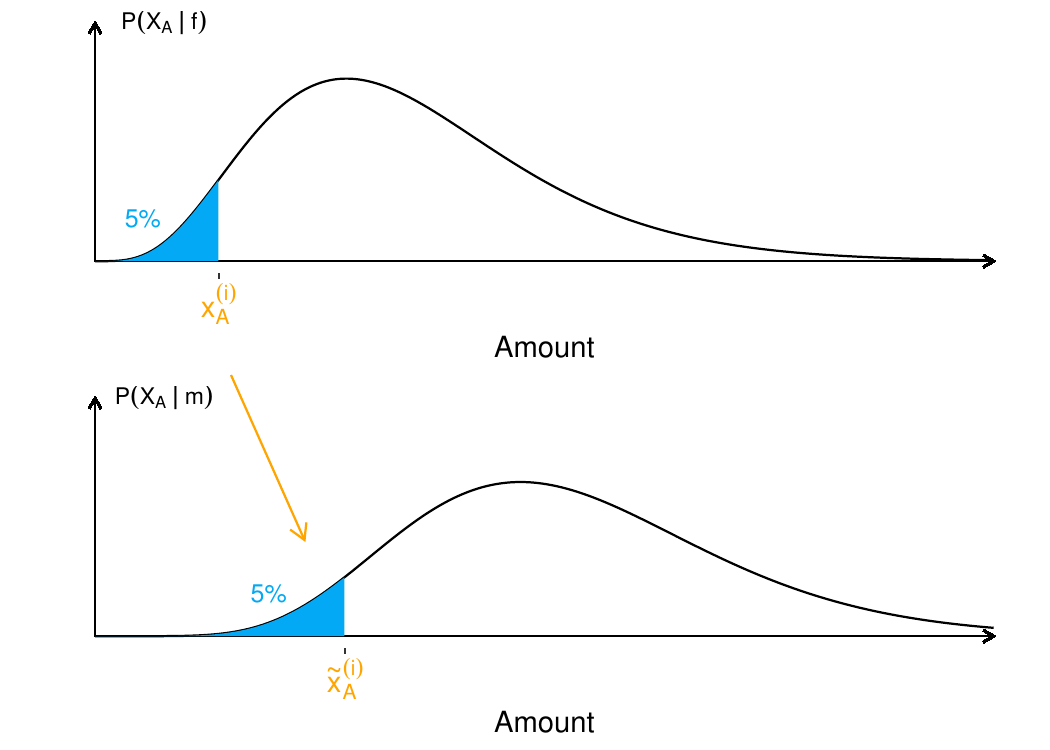}
    \caption{Rank-preserving interventional distribution: Transform female observation to the corresponding quantile in male distribution. }
    \label{fig:mod_int_distr}
\end{figure}

This intervention leads to \myemph{our estimand}; that is, the joint post-intervention distribution $P_p(G,C,X_A^{d_p}X_S^{d_p},Y^{d_p})$ in which the dashed arrows have been removed; thus, no effect of Gender on the mediators and the outcome exists -- but the distributions of males and females are comparable and still in line with the data-generating process on which we want to train our ML model. Additionally, our suggested intervention is ``rank-preserving'' in the sense that the quantile of female customers within their strata is transported into the FiND world (see Figure \ref{fig:mod_int_distr}).
Thereby, all relevant PA-dependent quantities %$(X_S, X_A, Y)$ 
are transformed into their FiND-world counterparts. 
%$(\tilde{x}_S^{(i)}, \tilde{x}_A^{(i)}, \tilde{y}^{(i)}) \ \forall i \in I_f$, where $I_f$ denotes the index set of female individuals.\footnote{Note that we do not modify male values since we assume that they are equal in both the real and the FiND world.}
Note that we can factorize the joint post-intervention distribution in line with the pre-intervention factorization (Eq.~\ref{formula:pre_int_fact}), but where the mediators and outcomes are replaced in line with the proposed intervention scheme. This leads to a $g$-formula type of factorization, which we can use for plug-in estimation of the relevant counterfactual distributions. A similar, quantile-based approach, can be found earlier in \citeA{plecko_fair_2020} which uses quantile regression forests for estimation. 

At this point, we assume that standard identification assumptions are met, such as conditional exchangeability. The latter is achieved by including all those variables in the adjustment set which guarantee that the back-door criterion is met \shortcite{pearl_causality_2009}. In our illustrative example, this implies that we assume that common causes between those variables that lie on causal pathways between the PA and outcome (i.e., mediators), and the target are measured. When using our framework in practice, identification assumptions should be checked carefully.

Note that it can be argued that multiple -- equally valid -- FiND worlds can be conceived: The concrete definition depends on the reference PA value used for the above intervention. Here we used the male class as reference PA value, but we could also use the female class or something in between as reference or baseline. All these choices are equally valid from the perspective of removing the gender effect and hence from a fairness perspective. While we use the male class for the explanations in the remainder of this section, we compare using the female class in the experiments in Section \ref{sec:experiments}, showing that the main difference is a general level shift in the predicted target values (for all observations) and some efficiency can be gained by exploiting data imbalance, if present.

\subsection{Estimation}
\label{sec:training}

We base our estimation algorithm on the factorization derived above, i.e.,  we use the empirical distributions of both $G$ and $C$ as a basis to implement the intervention, i.e., to obtain the intervention values from the quantiles of the respective post-intervention distributions of $X_A$, $X_S$, and $Y$. To determine the distributions and quantiles needed to facilitate the intervention implementation, our proposed algorithm uses the empirical distributions for the PA reference group (i.e., male customers) and a residual-based approach for the non-reference group (i.e., female customers). %Alternatively, we could data-adaptively estimate the quantiles from the conditional distributions \shortcite{hejazi_efficient_2022}, but we do not pursue this more complicated approach further in this manuscript. 
%Since we identified all causal quantities above with a statistical estimand, we can now turn to estimate the respective values, continuing with the same example.

\begin{figure}[ht]
        \centering
        \includegraphics[width=\textwidth, trim=10 180 10 180, clip]{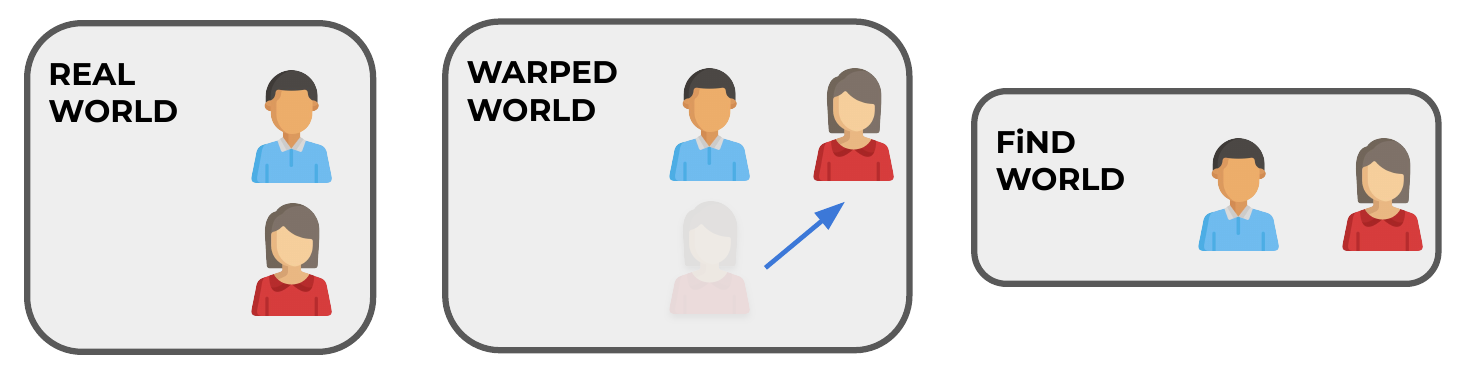}
        \caption{Illustration of the three worlds: 
        Real world discriminates between males and females, FiND world does not discriminate, warped world estimates FiND world by warping females to male level.}
        \label{fig:warped_world}
\end{figure}

More generally, we approximate the FiND world by ``warping'' the target and the features affected by the PA (see Figure \ref{fig:warped_world}). 
Once a preprocessed dataset containing ``cleaned'' features and target variables is available, standard ML techniques can be applied, prioritizing high predictive performance. 
This approach does not necessitate the incorporation of ``classical fairness metrics'' such as demographic parity in the training process.\footnote{For a more in-depth exploration of the philosophical rationale and the potential implications regarding the introduction of unfairness to the ADM system utilizing the trained ML model, please refer to \shortciteA{bothmann_what_2024}.}
The three key steps of our proposed algorithm are:

\begin{enumerate}
%    \item Derive or define the DAGs in the real world and in the FiND world.
    % \item Estimate the causal effects of the PA in the real world.
    \item Derive a warping from the real world to the warped world (see Section \ref{sec:warping}).
    %(approximating the FiND world)
    
    \item Train and test an ML model using the warped data (see Section \ref{sec:training_warped}).
    
    \item At the time of prediction, warp new data and obtain target predictions from trained ML model (see Section \ref{sec:prediction}).
\end{enumerate}

\subsubsection{Warping for Approximating the FiND World}
\label{sec:warping}
We propose to implement the interventions defined above (see Eq.\ \ref{eq:intervention}) by the following residual-based estimation method.
%For determining the male intervention values, we use the respective empirical conditional distributions, i.e., we use the actual measured values.
For determining the female intervention values, we must estimate --
for each variable to be warped -- 
(i) the individual probability rank of female $i$ (e.g., $p_A^{(i)}$ for variable $X_A$) and (ii) the corresponding quantile of the male distribution (e.g., $\txai$).
This means that we must estimate full distributions (not just location parameters) of $X_A|C=c, G=g$ for all values of $c$ and $g$ (analogously for $X_S$ and $Y$), which becomes prohibitively complex in situations with finite data and numeric confounders $C$ or features $X_*$. In our algorithm proposed below, we reduce estimation complexity by only estimating models for the location parameters of these distributions and derive individual probability ranks by using residuals of those models.
A computationally more complex alternative would be, e.g., to data-adaptively estimate the quantiles from the conditional distributions using the highly-adaptive LASSO \shortcite{hejazi_efficient_2022}.
% \textcolor{orange}{
% We propose to do this via the following steps (where this applies analogously to $X_S$ and $Y$, respectively):
% \begin{enumerate}
%     \item Estimate a prediction model $\pifem_A(C)$ for $X_A$ in the female population.
%     \item Derive individual probability ranks $p_A^{(i)}$ via residuals w.r.t.\ model $\pifem_A(C)$.
%     \item Estimate a prediction model $\pim_A(C)$ for $X_A$ in the male population.
%     \item Set $\hat{x}_A^{(i)} = \pim_A(c^{(i)}) + \tilde{q}_A^{(i)}$, where $\tilde{q}_A^{(i)} = \tilde{Q}_A(p_A^{(i)}|C = c^{(i)})$ is the $p_A^{(i)}$-quantile of the residuals of the male model $\pim_A(C)$.
% \end{enumerate}
% }
Figure \ref{fig:warping_vis_dig} visualizes our approach.
The five steps of this warping algorithm are explained for feature Amount ($X_A$), warping for other variables works analogously:

\begin{figure}[ht]
    \centering
    \includegraphics[width=\textwidth, trim=40 60 60 70, clip]{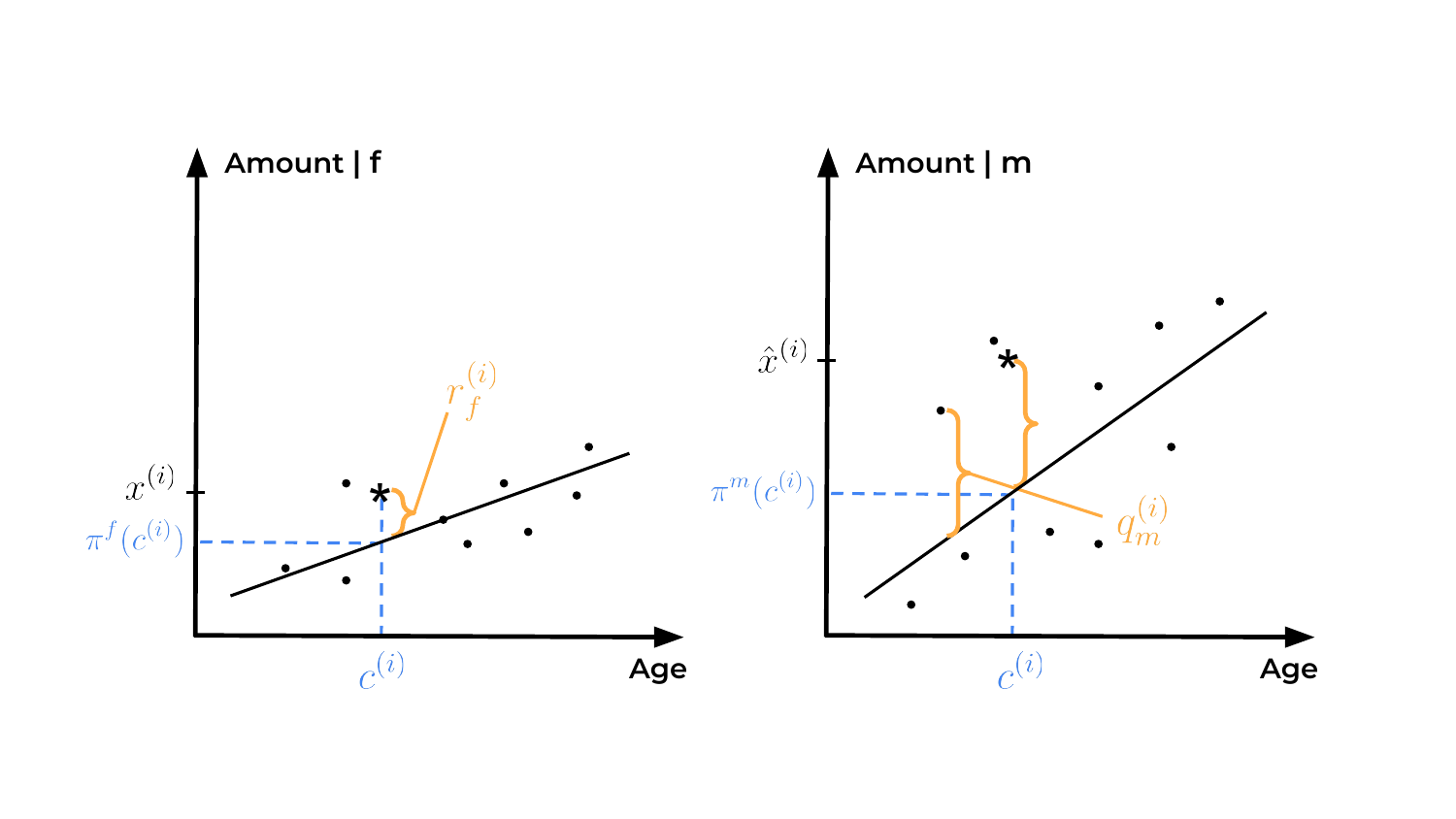}
    % https://docs.google.com/presentation/d/109Nh_x8RX8fgTJ0fpRoHHDckmIi_4KIzgw7wrjISYHs/edit#slide=id.p
    \caption{Visualization of residual-based warping approach: Real-world Amount of female individual (* in left plot) is warped to the respective quantile of male distribution (* in right plot). Subscript \textit{A} in $\pifem_A(c^{(i)}), x_A^{(i)}$, etc. is omitted for better readability.}
    \label{fig:warping_vis_dig}
\end{figure}

\noindent (1) Estimate prediction models $\pifem_A(C)$ for the female and $\pim_A(C)$ for the male population, where we are agnostic on the model class and can choose any ML model since we only rely on point predictions and model residuals on training data.
%assume that $X_A|C,G=g$ follows a Gamma-distribution (but the method is in general agnostic on this and we could equally well choose any other ML model since we only rely on point predictions and model residuals on training data), i.e.,
%$$ X_A|C, G=g \sim \text{Ga}(\alpha_g, \beta_g).$$

\noindent (2) Compute residuals as
\begin{align*}
r_f^{(i)} &= x_A^{(i)} - \pifem_A(c^{(i)}) \ \forall i \in I_f, \\
r_m^{(i)} &= x_A^{(i)} - \pim_A(c^{(i)}) \ \forall i \in I_m,
\end{align*}
where $I_f$ and $I_m$ are the female and male index sets, respectively.

\noindent (3) Compute the individual probability rank of female $i$ as ranked within the female residuals, i.e., telling us how ``exceptionally high or low'' her value is in comparison to other females, by
$$p_f^{(i)} = \frac{|\{j \in I_f: r_f^{(j)}\leq r_f^{(i)}\}|}{|I_f|} \quad \forall i \in I_f.$$

%(3) Analogously, after estimating parameters $\alpha_m$ and $\beta_m$, we compute the male residuals as
%$$r_m^{(i)} = \pim_A(c^{(i)}) - x_A^{(i)} \ \forall i \in I_m,$$
%where $I_m$ denotes the index set of male individuals.

\noindent (4) Set  $q_m^{(i)}$ to the empirical $p_f^{(i)}$-quantile of the residuals of the male model $\pim_A$, i.e.,
%$$\tilde{q}_A^{(i)} = \min \{r \in R_m: \text{at least} \ p_A^{(i)} \cdot 100 \% \text{ of } R_m \text{ are } \leq r \},$$
$$q_m^{(i)} = \min \{r \in R_m: \frac{|\{j \in R_m: j \leq r\}|}{|R_m|} \geq p_f^{(i)}\}\quad \forall i \in I_f,$$
where $R_m = \{r_m^{(i)}: i \in I_m\}$ is the set of male residuals.

\noindent (5) Finally, warp $x_A^{(i)}$ to $\hat{x}_A^{(i)}$ which is the sum of male prediction and warped residual, i.e.,
\begin{equation*}
    \hat{x}_A^{(i)} = \pim_A(c^{(i)}) + q_m^{(i)} \quad \forall i \in I_f.
\end{equation*}
Analogously, we can warp $X_S$ and $Y$ (where in the latter case, warped values of Amount and Savings must be plugged into the male prediction in step (5)). However, note that for warping of non-continuous variables (such as Savings and Risk), we define the models to predict the probability scores, not the hard labels. That way, the warped values, e.g., $\hat{x}_S^{(i)}$, are no longer binary, but may be $\in [-1, 2]$. If we need hard labels -- e.g., for learning a binary prediction model, such as for the target variable $Y$ in Section \ref{sec:training_warped} -- we can simply threshold these scores. On the other hand, for use in further warping steps (such as for warping of $\yi$), we can directly use the raw values by plugging them into the prediction function, thereby pulling through finer information than if we would threshold earlier in the process.

Now, we have warped all Gender-dependent quantities  $(x_S^{(i)}, x_A^{(i)}, y^{(i)})$ of female individuals to their warped world counterparts $(\hat{x}_S^{(i)}, \hat{x}_A^{(i)}, \hat{y}^{(i)})$, approximating their FiND world counterparts $(\tilde{x}_S^{(i)}, \tilde{x}_A^{(i)}, \tilde{y}^{(i)}) \ \forall i \in I_f$. To have a complete warped world dataset $\D_w = \Dsetw$, we set warped male values and values of non-warped features (e.g., Age) to their real-world value.
Additionally to having warped the training data, we have also estimated warping functions that can be applied for new test data at the time of prediction. %, see Section \ref{sec:prediction}.

\subsubsection{Training ML Models in the Warped World}
\label{sec:training_warped}

We can now use the warped world data $\D_w$ to train a prediction model for the warped target $\hat{Y}$. Assuming that the warping cleaned the data from any PA discrimination, we do not have to account for any fairness metrics in this training step but can just focus on training a model that has high predictive performance. Since we assume that all Gender-related discrimination was eliminated through the warping, we do not use Gender $G$ as a feature in this model (see Section \ref{sec:sim-results} for an investigation of what happens if this assumption is wrong, e.g., due to a misspecified DAG). 
%The usual steps of defining a performance metric (such as accuracy, F1-score, AUC, etc.) and choosing a resampling strategy apply. 
As a result, we obtain a trained model $f(\hat{\xv})$ which can be used for prediction. %in Section \ref{sec:prediction}.

\subsection{Prediction}
\label{sec:prediction}

Predicting new observations is a two-step process: First, the feature vector $\xv^*$ has to be warped, then the ML model is applied to the warped feature vector $\hat{\xv}^*$.

\paragraph{Warp New Data.} Consider a new observation $\xv^* =(g^*, x_A^*, x_S^*, c^*)$. If this is a male observation, no warping must be done; if this is a female observation, we use the estimated warping functions of Section \ref{sec:warping} as follows for $X_A$ and analogously for $X_S$ (but not for $Y$): %, since we do not have a value here -- which is why we developed a prediction model in the first place):

(1) Compute individual residual $r_f^*$ w.r.t.\ female model $\pifem_A(c^*)$ as $r_f^* = x_A^* - \pifem_A(c^*) $.

(2) Compute individual probability rank $p_f^*$ w.r.t.\ female population $I_f$ as above. % of training data

%$$p_A^* = \frac{|\{j \in I_f: r_f^{(j)}\leq r_f^*|\}}{|I_f|}.$$

(3) Set $q_m^*$ to the empirical $p_f^*$-quantile of training data residuals of male model $\pim_A$ as above.

%$$\tilde{q}_A^* = \min \{r \in R_m: \text{at least} \ p_A^* \cdot 100 \% \text{ of } R_m \text{ are } \leq r \}.$$

(4) Warp $x_A^*$ to the sum of male prediction and warped residual, i.e., $\hat{x}_A^* = \pim_A(c^*) + q_m^*$.

\noindent After carrying out the same steps for warping $X_S$, we finally obtain the warped observation $\hat{\xv}^* =(\hat{x}_A^*, \hat{x}_S^*, c^*)$ (recall that we do not use Gender as a feature in the prediction model).

\paragraph{Predict New Data.}
%\label{sec:pred_new}
For predicting the target in the warped world $\hat{y}^*$, we plug the warped observation $\hat{\xv}^*$ into the prediction model trained on the warped world data, i.e., $\hat{y}^* = f(\hat{\xv}^*)$.

\section{Evaluation}
\label{sec:evaluation}

We propose evaluation criteria that can be used for two purposes: Section \ref{sec:eval-method} describes how to evaluate our proposed warping method for rank-preserving interventional distributions in a simulation study. Section \ref{sec:eval-usecase} describes how the warped data and resulting ML models can be evaluated in an applied use case. We denote with $\pihi$, $\pithi$, and $\psihi$ the predicted target of individual $i$ in the real, warped, and FiND world, respectively.

\subsection{Evaluation of Warping Method}
\label{sec:eval-method}

We can evaluate our warping method w.r.t.\ (i) the warped data, asking, e.g., if the FiND world is recovered by the warping and w.r.t.\ (ii) the final ML model -- using the warped data.
\paragraph{(W1) Recovering FiND world.}
In a simulation study %(knowing the distributions in the FiND world) 
we can compare the warped and the FiND world distributions to investigate if the warping procedure recovers the FiND world (and thus the estimation procedure is unbiased, as identification assumptions are met by construction in our setup). For numerical features, we compare warped world and FiND world empirical distributions by Kolmogorov-Smirnov (KS) tests \shortcite{smirnov_estimate_1939}, and for binary features, we use binomial tests \shortcite{clopper_use_1934}. 
Additionally, we use a t-test \shortcite{helmert_genauigkeit_1876} to test the null hypothesis that there is no discrimination in the warped world between male and female subgroups w.r.t.\ risk predictions.
If the method works, the p-values of these tests should be consistently high, indicating that there is strong support in favor of the null hypotheses.

\paragraph{(W2) Identifying strongest discriminated individuals.}
In addition to the population-wide perspective of (W1), we are interested in the individual perspective, i.e., if the warping method also recovers the individual ranks of the FiND world w.r.t.\ the target variable prediction. If this were the case, we could identify individuals who are most strongly affected by discrimination in the real world by comparing real-world and warped-world predictions in an applied use case.
For the warped class of the PA, we compute individual risk prediction differences between the real world and the warped world, $d_1^{(i)} = \pihi - \pithi$ and between the real world and the FiND world, $d_2^{(i)} = \pihi - \psihi$, respectively, after training ML models on the respective datasets.
%Strongest negatively affected individuals are at the upper end of the distribution. As can be seen, top discriminated individuals (high diff between real and FiND world) are correctly identified (high diff between real and warped world).
We use a t-test to test the null hypothesis that the means of these differences ($d_1$ and $d_2$) are equal, i.e., $H_0: d_1 = d_2$. If the method works, p-values of these tests should be consistently high, and differences $d_1^{(i)} - d_2^{(i)}$ should be small.
Correlation between ranks of $d_1^{(i)}$ and $d_2^{(i)}$ should be high, too.

\subsection{Evaluation in an Applied Use-Case}
\label{sec:eval-usecase}

How can the model be evaluated in an applied use case, i.e., how can we know if the warping method worked and if it removed unfairness? In our opinion, this cannot be answered by evaluating the final ML model w.r.t.\ some ``classical'' fairML metrics.\footnote{These kinds of metrics (such as demographic parity, equalized odds, etc.) do not reflect a clearly defined concept of fairness and, hence, are not suitable for deciding if an ML model entails unfairness. However, as readers might still be interested in the respective values -- in the sense of ``fairness-related performance metrics'' which allow insights into the predictive performance of the resulting ML model -- we provide the resulting metrics in Appendix \ref{sec:app_classical_fairml}.} Once we have successfully warped the data from the real to the warped world (approximating the FiND world), we reduced the problem to finding a model with good predictive performance.
However, we can train models in the real and the warped world and then compare their behavior:

\paragraph{(UC1) Comparing performance.}
Test performance of the ML models in the real world $\pihf$ and the warped world $\pithf$ can be compared, assuming that both models fit ``their'' world equally well. However, this %is just a qualitative comparison and 
must not be misinterpreted as either of those models being better than the other one, as the models are merely modeling different worlds. % For simulated data, we can additionally evaluate if the performance in the warped and FiND world are similar -- which they should be if the warping worked.

\paragraph{(UC2) Comparing predictions and identifying strongest discriminated individuals.}
For each individual $i$, the predictions in the real and the warped worlds can be compared by computing the difference $d_1^{(i)}$. As for (W2), this analysis can reveal individuals who are discriminated most in the real world (either positively or negatively). Additionally, these differences can be aggregated on the subgroup level, and tests can be computed to test the null hypothesis that predictions do not change between the two worlds for the respective subgroup. %For simulated data, we can additionally test if predictions in the warped and FiND world are similar -- which they should be if the warping worked.

\paragraph{(UC3) Identifying strongest warped individuals.}
We can also ask which individuals are affected most by the warping. These individuals' feature vectors have the largest distance between the real and the warped world, i.e., %, meaning that the respective individuals are discriminated most in the real world (either positively or negatively).
we can compare $\xi$ and $\hat{\xv}^{(i)}$ by a suitable distance metric for each individual $i \in \nset$. %in several ways. For example, if there are categorical features, we can utilize the Gower distance. Otherwise, we can use the Euclidean distance, leading to a distance $d_0^{(i)}$ for each individual $i \in \nset$. %For simulated data, we can additionally evaluate if this detection reveals correctly the observations that differ most between real and FiND world.

\paragraph{(UC4) Identifying important features.}
For each feature, we can compare the empirical distributions in the real and the warped world, % (and (iii) the FiND world for simulated data),
i.e., of $X_j$ and $\hat{X}_j$ for each $j \in \pset$. We compute distances for each (normalized) feature and, thereby, can identify features that vary most between the two worlds, which can give us an indication of their relevance in terms of discrimination w.r.t.\ the PA. % For a real-world use case, we can detect features that are affected most by the mapping and in which sense. %For simulated data, we can additionally evaluate if this detection reveals correctly the features that differ most between real and FiND world.

\section{Experiments}
\label{sec:experiments}

To investigate the behavior of our proposed method RPID, we first conduct a simulation study where we know the true DAG in both the real and the FiND world in Sections \ref{sec:sim} and \ref{sec:sim-results}. 
Subsequently, we apply the methods to the German Credit dataset \cite{hofmann_statlog_1994} in Section \ref{sec:results_german}. Additionally to our proposed method RPID, we investigate \textit{fairadapt} \cite{plecko_fair_2020} and compare results. %\footnote{Note that we do not compare with \textit{fairadapt} on the simulated data because the philosophical idea of \textit{fairadapt} is slightly different than that of RPID and the research questions formulated below in the simulation study are not directly meaningful for \textit{fairadapt}.}

\subsection{Simulation Study Setup}
\label{sec:sim}

We seek to answer the following \textbf{research questions}: 

\begin{enumerate}[label=(RQ\arabic*)]
    \item Does our warping method work as expected? In other words, does this method recover the distributions in the FiND world (W1), and can it correctly identify the individual ranks w.r.t.\ the target in the FiND world (W2)?
    \item How does misspecification of the DAG affect the results?
    \item What effects does the direction of warping have on performance (e.g., if subgroup A of the PA is warped to subgroup B, versus the other way around)?
\end{enumerate}

\paragraph{Data simulation setup.}
We simulate data from the DAGs depicted in Figure \ref{fig:DAG}, using the R package \texttt{simcausal} \shortcite{sofrygin_simcausal_2017}. The real-world data simulation contains all arrows, while the FiND world data simulation only contains solid arrows by setting Amount, Savings, and Risk of females to their corresponding values among the male distributions.
%Gender to \textit{male} -- in the conditional distributions of the descendants of Gender -- for all observations, thereby eliminating the Gender effect. 
The distributions used here are (left: real-world, right: FiND world):\footnote{Concrete values can be found in \texttt{simulation\_study.R} in the following GitHub repo: \url{https://github.com/slds-lmu/paper\_2023\_cfml}}
%\texttt{simulation\_study.R} in \url{https://github.com/slds-lmu/paper\_2023\_cfml}}
\begin{align*}
    G &\sim \text{B}(\pi_G) & G &\sim \text{B}(\pi_G)\\
    C &\sim \text{Ga}(\alpha_C, \beta_C) & C &\sim \text{Ga}(\alpha_C, \beta_C)\\
    X_A|C, G &\sim \text{Ga}(\alpha_A(C, G), \beta_A(C,G))& \tilde{X}_A|C &\sim \text{Ga}(\alpha_A(C, m), \beta_A(C,m))\\
    X_S|C, G &\sim \text{B}(\pi_S(C,G))& \tilde{X}_S|C &\sim \text{B}(\pi_S(C,m))\\
    Y|X_A, X_S, C, G &\sim \text{B}(\pi_Y(X_A, X_S, C, G))& \tilde{Y}|\tilde{X}_A, \tilde{X}_S, C &\sim \text{B}(\pi_Y(\tilde{X}_A, \tilde{X}_S, C, m)),
\end{align*}
\noindent where we use linear combinations of the features combined with a log- and logit-link for the Gamma and Binomial models, respectively, and mirror the Gender distribution of the German Credit data with $\pi_f=31\%$ females. We perform $M=1,000$ simulations on datasets of size $N_{tr}=10,000$ for training and of size $N_{te}=1,000$ for test, for each world, using the same seed for the two worlds to ensure comparability. Note that Gender and Age are then identical in both worlds, and only the descendants of Gender have differing values. We refer to this setup as (SIM1).
To answer the misspecification behavior question (RQ2), we modify the simulation slightly by assuming a different DAG, i.e., a DAG in which Gender does affect Age at credit application time. We reflect this assumed structure by sampling Age from the data-generating mechanism via $C \sim \text{Ga}(\alpha_C(G), \beta_C(G))$, i.e., from different distributions for female and male observations.
However, we ignore this additional arrow in the DAG for warping, i.e., the warping assumptions do not mirror the data-generating process correctly and we do not warp Age. 
We refer to this setup as (SIM2).

\paragraph{Warping and prediction models.}
For warping models, we estimate models following the same distributional assumptions as in the simulation, i.e., by estimating the parameter vectors of the Gamma and logistic regressions %$\alpha_A, \beta_A, \pi_S, \pi_Y$,
separately for male and female observations of the training data. With these models, we apply the above warping strategy.
As prediction models for the target variable, we train logistic regression models on the training data, warped training data, and FiND world training data, separately. 

\paragraph{Comparison with fairadapt.} We compare our results with the results obtained by using \textit{fairadapt} \cite{plecko_fair_2020}, which is also a preprocessing method aiming at adapting real-world data. They aim at computing \textit{fair twins} for each observation: The PA of a given observation is changed to the baseline value of the PA by a \textit{do}-intervention. This is different from our proposal (Eq. \ref{eq:intervention}), which does not require the \shortcite<philosophically criticized, see, e.g.,>{issa_kohler-hausmann_eddie_2019} conceptualization of interventions on the PA, e.g., gender or race. Subsequently, the descendants of the PA are transformed to conserve individual merits (using quantiles, similar to our proposal, but in a different counterfactual world). Overall, while there are similarities between the two approaches, \citeA{plecko_fair_2020} do not define a FiND world comparably to ours and rather aim at minimizing ``the distortion in the data coming from the projection''. They adopt a similar view as \textit{counterfactual fairness} \cite{kusner_counterfactual_2017} by aiming at equal predictions for fair twins. %(see \cite{anonymous_2024}, Section 3.3.2, for a detailed comparison of the philosophical differences between our approach and counterfactual fairness). 
As an estimation method they use quantile regression forests \cite{meinshausen_quantile_2006}. We will refer to their world of fairness-adapted twins as \textit{adapt world}. We used their implementation in the R package \texttt{fairadapt} with default hyperparameters.\footnote{\url{https://cran.r-project.org/web/packages/fairadapt/index.html}}

\subsection{Simulation Study Results}
\label{sec:sim-results}

%\paragraph{Research questions.}
With these models, we can now answer the above research questions, using a significance threshold of $\alpha=5\%$ for all tests. All reported results are on test data.

\subsubsection{Results -- RQ1}

In this section, we investigate if FiND world distributions are recovered by the warping methods (W1) and if the strongest discriminated individuals can be identified (W2).

\begin{figure}[ht]
    \centering
    \includegraphics[width=0.49\textwidth]{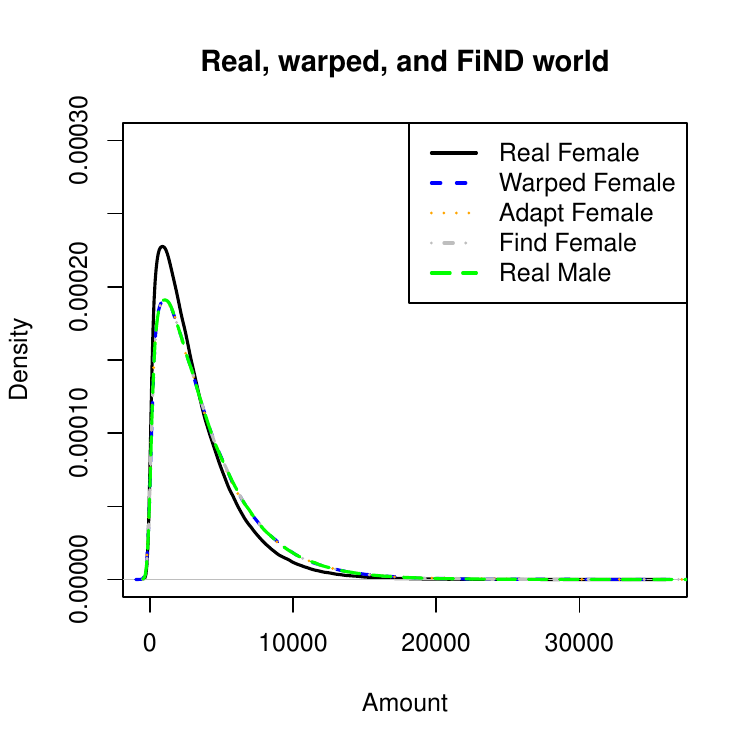}
    \caption{Average distribution of Amount in different worlds.}
    \label{fig:rq1-1}
\end{figure}

\paragraph{(W1) Recovering FiND world -- RPID:} Figure \ref{fig:rq1-1} shows the distribution of Amount $X_A$ in the different worlds for male and female observations, aggregated over all iterations of the simulation study: Warping with RPID seems to recover the FiND world distribution very well. %The visual impression of similar female distributions in warped and FiND world is supported by a p-value of $99.9\%$ of the respective KS test. Figure \ref{fig:rq1-2} shows the distribution of p-values over all iterations for Amount, Saving, and Risk. %, where for the categorical variables, a $\chi^2$-test is used.
%Table XXX shows distributions of Saving and Risk for one iteration.
%We conclude that warping recovers the marginal distributions in the FiND world for each subgroup of the PA since the 
The null hypothesis of equal distributions in the warped and the FiND world is only rejected in $0\%, 0.4\%, 0\%$ of the iterations for Amount, Savings, and Risk, respectively.
The mean difference between male and female risk predictions in the real world is on average $0.1122 \ (2.5\%$ and $97.5\%$ quantiles in the $M=1,000$ simulation runs:  $(0.0973, 0.1275)$). In the warped world, this is reduced to $-0.0016$  $(-0.0153, 0.0120)$, meaning that the difference between subgroups varies very closely around zero, i.e., we effectively reduced PA discrimination. (In the FiND world, these values are exactly $0$ by construction.) 

\paragraph{(W1) Recovering FiND world -- fairadapt:}
As Figure \ref{fig:rq1-1} shows, also warping via \textit{fairadapt} seems to recover the FiND world distribution of Amount very well. The null hypothesis of equal distributions in the adapt and the FiND world is rejected in $8.8\%, 27.6\%,$ $17.9\%$ of the iterations for Amount, Savings, and Risk, respectively.
The mean difference between male and female risk predictions in the adapt world is on average $0.0008 \ (2.5\%$ and $97.5\%$ quantiles:  $(-0.0267, 0.0292)$). This means PA discrimination is also reduced by \textit{fairadapt} but not as much as by RPID.

\begin{figure}[ht]
    \begin{subfigure}[b]{0.49\textwidth}
         \centering
         \includegraphics[width=\textwidth]{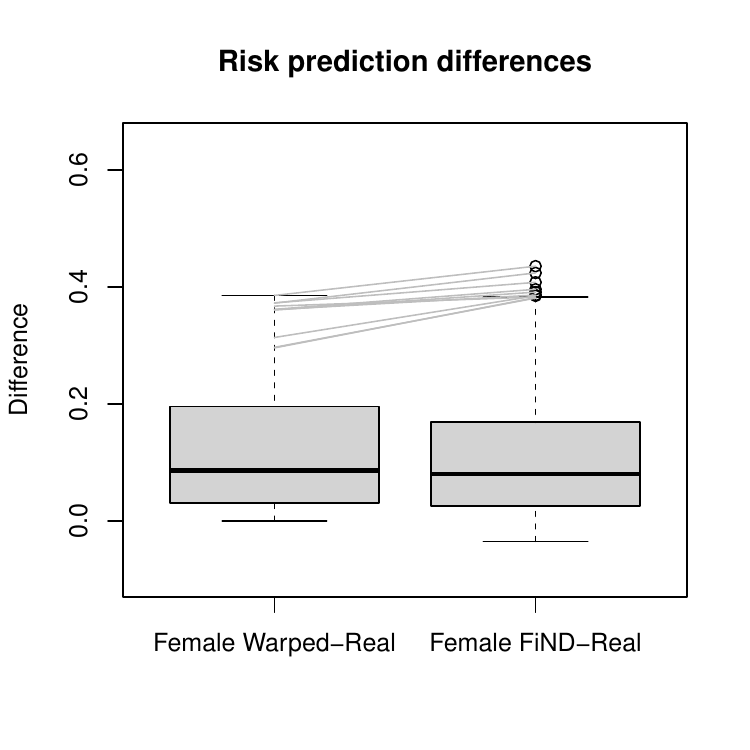}
         \caption{}
         \label{fig:rq1-2}
     \end{subfigure}
     \hfill
         \begin{subfigure}[b]{0.49\textwidth}
         \centering
         \includegraphics[width=\textwidth]{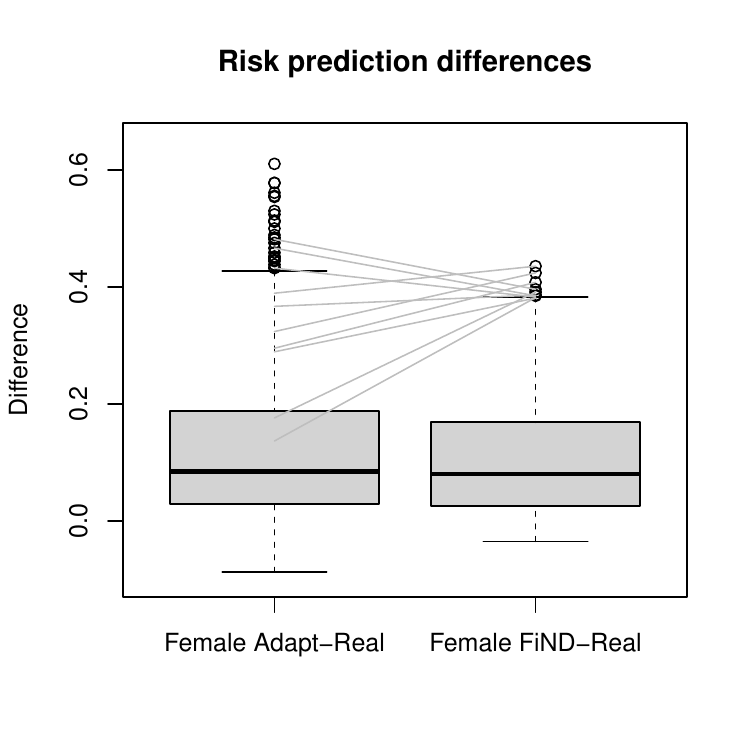}
         \caption{}
         \label{fig:rq1-2-adapt}
     \end{subfigure}
     \caption{Identification of strongest discriminated individuals with (a) RPID and (b) fairadapt for the same iteration of the simulation study.}
\end{figure}

\paragraph{(W2) Identifying strongest discriminated individuals -- RPID:}
Investigating individual predictions, we see that correlations between ranks in the warped and the FiND world are high (mean: $0.893$). % and significantly higher than using the baseline ($0.822, \ p<10^{-15}$). 
Figure \ref{fig:rq1-2} shows individual risk prediction differences between the real world and the warped world as well as between the real world and the FiND world for females in one iteration.
The most strongly negatively affected individuals are at the upper end of the distribution. As shown, the most discriminated individuals (large difference between the FiND world and the real world) are correctly identified (large difference between the warped world and the real world).
Looking at all simulation runs, in $86\%$ of iterations, the null hypothesis of equal mean differences $d_1$ (difference between real and warped world) and $d_2$ (difference between real and FiND world) cannot be rejected ($81\%$ in female subgroup). In cases with $p<0.05$, the mean difference is $-0.0003$ ($-0.0023$ in female subgroup) -- meaning that the deviation between the warped world and the FiND world is  minimal in these cases. %For the baseline, the null hypothesis is always rejected, with a mean difference of $-0.0456$.
The mean difference between $d_1$ and $d_2$ is on average $-0.0003 \ (2.5\%$ and $97.5\%$ quantiles: $(-0.0081, 0.0079)$ -- in female subgroup: $-0.0011 (-0.0154, 0.0133)$).
We conclude that warping (i) recovers the marginal distributions in the FiND world, (ii) effectively removes discrimination, and (iii) correctly identifies the most discriminated individuals. %, and (iv) outperforms the baseline.

\paragraph{(W2) Identifying strongest discriminated individuals -- fairadapt:}
Correlations between ranks in the adapt and the FiND world are slightly smaller (mean: $0.811$). Figure \ref{fig:rq1-2-adapt} shows individual risk prediction differences between the real world and the adapt world.
Most discriminated individuals are not identified as confidently as with RPID.
In $56\%$ of iterations, the null hypothesis of equal mean differences $d_1$ and $d_2$ cannot be rejected ($49\%$ in female subgroup). In cases with $p<0.05$, the mean difference is $0.0063$ ($0.0065$ in female subgroup) -- meaning that the deviation between the adapt world and the FiND world is also small in these cases, but considerably higher than for RPID.
The mean difference between $d_1$ and $d_2$ is on average $-0.0030 \ (2.5\%$ and $97.5\%$ quantiles: $(-0.0110, 0.0168)$ -- in female subgroup: $-0.0034 \ (-0.0242, 0.0309)$).
%The mean difference between $d_1$ and $d_2$ for females is on average $0.0034 \ (2.5\%$ and $97.5\%$ quantiles:  $(-0.0242, 0.0309)$).
We conclude that \textit{fairadapt} is also able to reduce discrimination, but performs worse than RPID regarding the identification of most discriminated individuals.

\subsubsection{Results -- RQ2}

In this section, we investigate how misspecification (differing Age distributions per Gender) affects the results.

\paragraph{RPID:}
The null hypothesis of equal distribution in the warped world and the FiND world is rejected in $17\%, 4\%, 0\%$ of the iterations for Amount, Savings, and Risk, respectively.
%are comparable to (RQ1), see Figure \ref{fig:rq4-1-2} (B) \lb{Check ratio of WRS-tests where the null is not rejected -- should be smaller than in RQ2} (C) \lb{Check same number for RQ3}
In the real world, the mean difference between female and male risk predictions is $0.1723 \ (0.1565, 0.1892$), which is higher than in (SIM1). In the warped world, this is reduced to $0.0355 \ (0.0196, 0.0509)$, i.e., by a factor of $4.9$, but does not vary around zero as above. 
The correlation of ranks (mean: $0.9518$) is higher than above since discrimination in the FiND world is higher in (SIM2).
In $61\%$ of iterations, the null hypothesis of equal mean differences $d_1$ and $d_2$ cannot be rejected ($6.9\%$ in female subgroup). In cases with $p<0.05$, the mean difference is $0.0107$ ($0.0255$ in female subgroup) -- far higher than above.
The mean difference between $d_1$ and $d_2$ is on average $0.0067 \ (2.5\%$ and $97.5\%$ quantiles: $(-0.0016, 0.0147)$ -- in female subgroup: $0.0244 \ (0.0089, 0.0398)$).
We conclude that misspecification of the DAG is a relevant factor for degrading the performance of our approach.

\paragraph{fairadapt:}
The null hypothesis of equal distribution in the adapt world and the FiND world is rejected in $12\%, 93\%, 90\%$ of the iterations for Amount, Savings, and Risk, respectively, indicating a rather poor performance in recovering the FiND world distributions.
The mean difference between female and male risk predictions is $0.0557 \ (0.0240, 0.0877)$, i.e., does not vary around zero as above and is slightly higher than for RPID. 
The correlation of ranks (mean: $0.8222$) is slightly higher than above but substantially smaller than for RPID.
% In $0.6\%$ of iterations, the null hypothesis of equal differences $d_1$ and $d_2$ cannot be rejected. In cases with $p<0.05$, the mean difference is $0.055$ -- higher as above and higher as for RPID.
% The mean difference between $d_1$ and $d_2$ for females is on average $0.0547 \ (2.5\%$ and $97.5\%$ quantiles:  $(0.0240, 0.0872)$).
In $1.4\%$ of iterations, the null hypothesis of equal mean differences $d_1$ and $d_2$ cannot be rejected ($0.6\%$ in female subgroup). In cases with $p<0.05$, the mean difference is $0.0271$ ($0.0550$ in female subgroup) -- also substantially higher than above.
The mean difference between $d_1$ and $d_2$ is on average $0.0268 \ (2.5\%$ and $97.5\%$ quantiles: $(0.0110, 0.0433)$ -- in female subgroup: $0.0547 \ (0.0240, 0.0872)$), i.e., does not vary around zero as above and is slightly higher than with RPID. 
We conclude that misspecification of the DAG is also a relevant factor for degrading the performance of the \textit{fairadapt} approach, even slightly worse than for RPID.

\subsubsection{Results -- RQ3} 

In this section, we investigate the effects of the warping direction, i.e., warping the larger class (\textit{male}) to the smaller class (\textit{female}).

\begin{figure}[ht]
     \centering
     \hfill
     \begin{subfigure}[b]{0.49\textwidth}
         \centering
         \includegraphics[width=\textwidth]{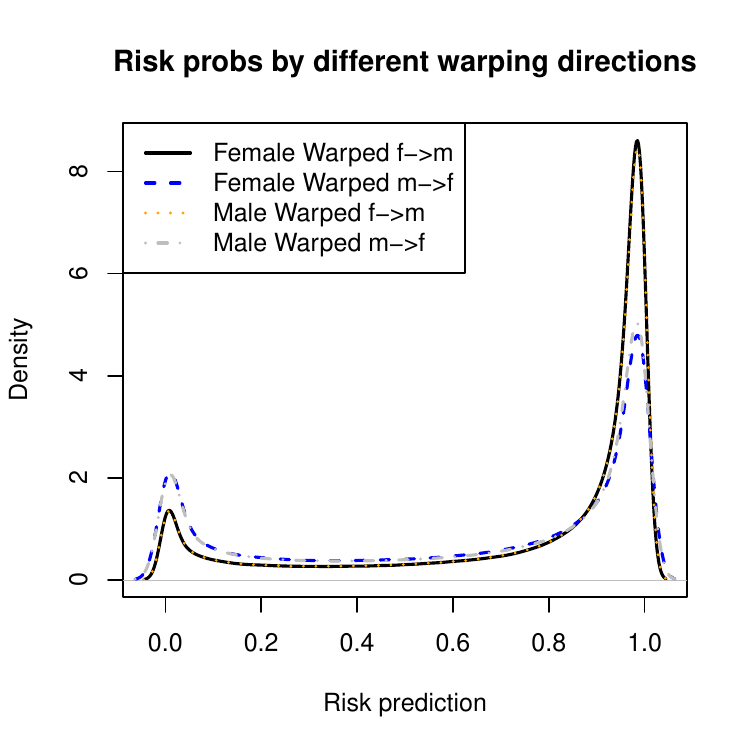}
         \caption{}
        \label{fig:rq5-general-level-shift}
     \end{subfigure}
     \hfill
     \begin{subfigure}[b]{0.49\textwidth}
         \centering
         \includegraphics[width=\textwidth]{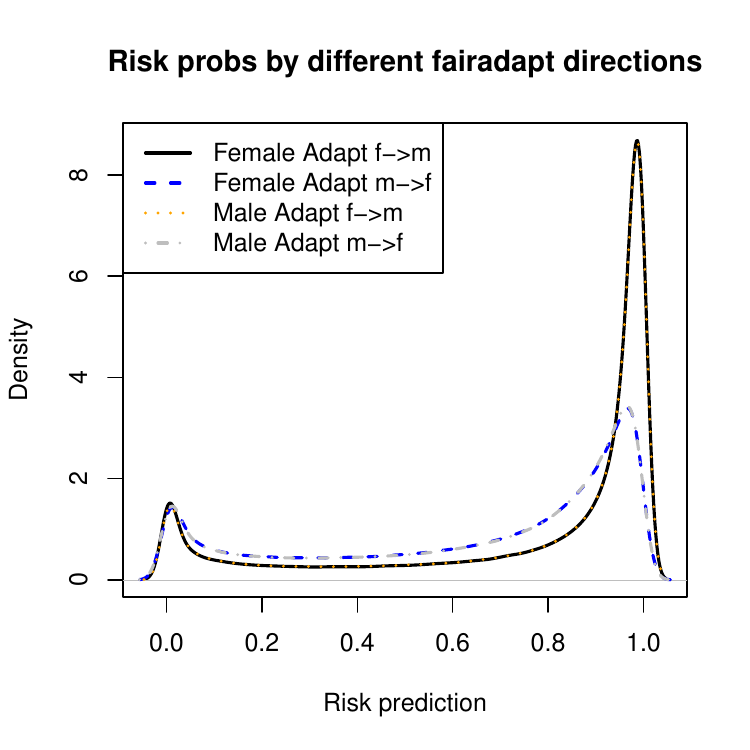}
         \caption{}
        \label{fig:rq5-general-level-shift-adapt}
     \end{subfigure}
        \caption{Risk predictions for different warping directions for (a) RPID and (b) fairadapt.}
        \label{fig:rq1}
\end{figure}

\paragraph{RPID:}
By switching the warping direction, we observe the following:
Recovering marginal FiND world distributions is equally successful as when warping female to male values. The null hypothesis is rejected in $0\%, 0.3\%, 0\%$ of the iterations for Amount, Savings, and Risk, respectively. % (see also Figure \ref{fig:rq5-1-1}).
The mean difference between risk predictions in the warped world is reduced to $0.0065 \ (-0.0104, 0.0260)$ -- slightly worse than in the analysis of RQ1, which is due to the imbalance of the data.
The mean correlation of ranks compared with ranks of RQ1 is high ($0.9595$), meaning that individual ranks are comparable for both warping directions.
In $64\%$ of iterations, the null hypothesis of equal mean differences $d_1$ and $d_2$ cannot be rejected ($34\%$ in female subgroup). In cases with $p<0.05$, the mean difference is $0.0039$ ($0.0073$ in female subgroup) -- meaning that the deviation between the warped world and the FiND world is also minimal in these cases (although a bit higher than in the analysis of RQ1, due to data imbalance).
The mean difference between $d_1$ and $d_2$ is on average $0.0017 \ (2.5\%$ and $97.5\%$ quantiles: $(-0.0115, 0.0149)$ -- in female subgroup: $0.0050 \ (-0.0077, 0.0188)$).
This means we can also mitigate discrimination and preserve individual ranks by changing the warping direction. However, the general level of the risk predictions changes, as shown in Figure \ref{fig:rq5-general-level-shift}, which had to be expected, since we are now warping male to female values.

\paragraph{fairadapt:}
By switching the direction and adapting male to female values, we observe the following:
Recovering marginal FiND world distributions is less successful as when adapting female to male values. The null hypothesis is rejected in $8.9\%, 26.5\%, 12.4\%$ of the iterations for Amount, Savings, and Risk, respectively. % (see also Figure \ref{fig:rq5-1-1}).
The mean difference between risk predictions in the adapt world is reduced to $0.0010 \ (-0.0262, 0.0288)$ -- comparably to the above.
The mean correlation of ranks compared with ranks of RQ1 is high ($0.9791$), meaning that individual ranks are comparable for both adapting directions.
% In $19\%$ of iterations, the null hypothesis of equal differences $d_1$ and $d_2$ cannot be rejected. In cases with $p<0.05$, the mean difference is $-0.0016$, meaning that the deviation between the adapt world and the FiND world is also minimal in these cases.
% The mean difference between $d_1$ and $d_2$ for females is on average $-0.0013 \ (2.5\%$ and $97.5\%$ quantiles:  $(-0.0150, 0.0115)$).
In $51\%$ of iterations, the null hypothesis of equal mean differences $d_1$ and $d_2$ cannot be rejected ($19\%$ in female subgroup). In cases with $p<0.05$, the mean difference is $-0.0033$ ($-0.0016$ in female subgroup).
The mean difference between $d_1$ and $d_2$ is on average $-0.0018 \ (2.5\%$ and $97.5\%$ quantiles: $(-0.0187, 0.0160)$ -- in female subgroup: $-0.0013 \ (-0.0150, 0.0115)$).
This means we can also mitigate discrimination and preserve individual ranks with changing the adapting direction. The general level of the risk predictions changes as for RPID, see Figure \ref{fig:rq5-general-level-shift-adapt}. 

\subsubsection{Discussion} 
\label{sec:discussion_sim}

We summarize our findings with respect to the research questions:

\begin{enumerate}[label=(RQ\arabic*)]
    \item We have seen that our warping approach recovers the marginal distributions in the FiND world very well. It effectively removes PA-related discrimination and correctly identifies the most discriminated individuals. The \textit{fairadapt} approach is also able to reduce discrimination, but to a smaller degree, and can also detect discriminated individuals, but less effectively than RPID.
    \item Misspecification of the DAG is a relevant challenge for both RPID and \textit{fairadapt}, where RPID seems to be slightly more robust in the investigated simulation setup than \textit{fairadapt}.
    \item The direction of warping has no relevant impact on the performance of both approaches. The effects that could be seen above are merely due to imbalance in the PA Gender. We therefore recommend warping (or adapting) the smaller subgroup towards the larger subgroup.
\end{enumerate}

Overall, this means that RPID is a promising approach that can be applied for at least two goals: (i) When aiming at mitigating PA-related discrimination in an ADM system, RPID can be used to de-bias the real-world data, such that a PA-neutral ML model can be trained in the warped world. (ii) When aiming at diagnosing individual PA-related discrimination in the real world, RPID can be used to estimate the difference between risk predictions in real and FiND world. A thereby derived score describes real-world discrimination of a given individual and can also be used to compare the degree of (positive or negative) discrimination of different individuals. It can also be analyzed on the subgroup level after suitable aggregations.

% We have shown that for the simulation setup above, our proposed method works as expected, recovering the marginal distributions in the FiND world and individual ranks;  %misspecification of the DAG degrades performance and 
% direction of warping does not make a relevant difference.
% %We have also shown how to use the method in an applied use case. 
However, as this is just an initial study, these investigations should be extended by follow-up work. As subsequent investigations, we would propose to (at least): (i) consider other, diverse DAGs, (ii) compare different ML models for warping and target prediction, and (iii) investigate behavior on other empirical datasets.

%A general limitation of our method is that it depends on knowing the true DAG. 
%As shown in RQ2, misspecifying the DAG degrades the performance of the method. 
Developing an SCM that captures the actual data-generating process can be challenging. In applications where knowledge about human decisions is available (e.g., medical decisions, banking decisions), this may be possible, but it is more challenging in settings where complex biological or psychological processes are involved.
Hence, special care should be given to constructing the SCM in an applied use case by carefully interweaving expert knowledge on the subject matter and rigorous application of causal discovery methods \cite<see, e.g.,>{nogueira_methods_2022}.

\subsection{German Credit Data}
\label{sec:results_german}

We assume the same DAG as in the simulation study, depicted in Figure \ref{fig:DAG}. For warping and prediction models, we use the same models as in the simulation study.

For the evaluation of the behavior of our method for this applied use case, we use the evaluation strategies defined in Section \ref{sec:eval-usecase}. Models are trained on randomly sampled $80\%$ of the training data (i.e., 800 from 1,000 observations) and tested on the remaining $20\%$. %We apply the four evaluation strategies from Section \ref{sec:eval-usecase}.

\subsubsection{Comparing performance -- UC1}
\paragraph{RPID:} Accuracy on the test set in the real world is $71\%$ for both the male and the female subgroup. In the warped world, male accuracy is comparable, with test accuracy of $70\%$. However, female accuracy increases to $75\%$, showing increasing performance for the discriminated subgroup.

\paragraph{fairadapt:} \textit{fairadapt} achieves $72\%$ and $69\%$ test accuracy for males and females, respectively.

\subsubsection{Comparing predictions and identifying strongest discriminated individuals -- UC2}

\begin{figure}[ht]
     \centering
     \begin{subfigure}[b]{0.49\textwidth}
         \centering
         \includegraphics[width=\textwidth]{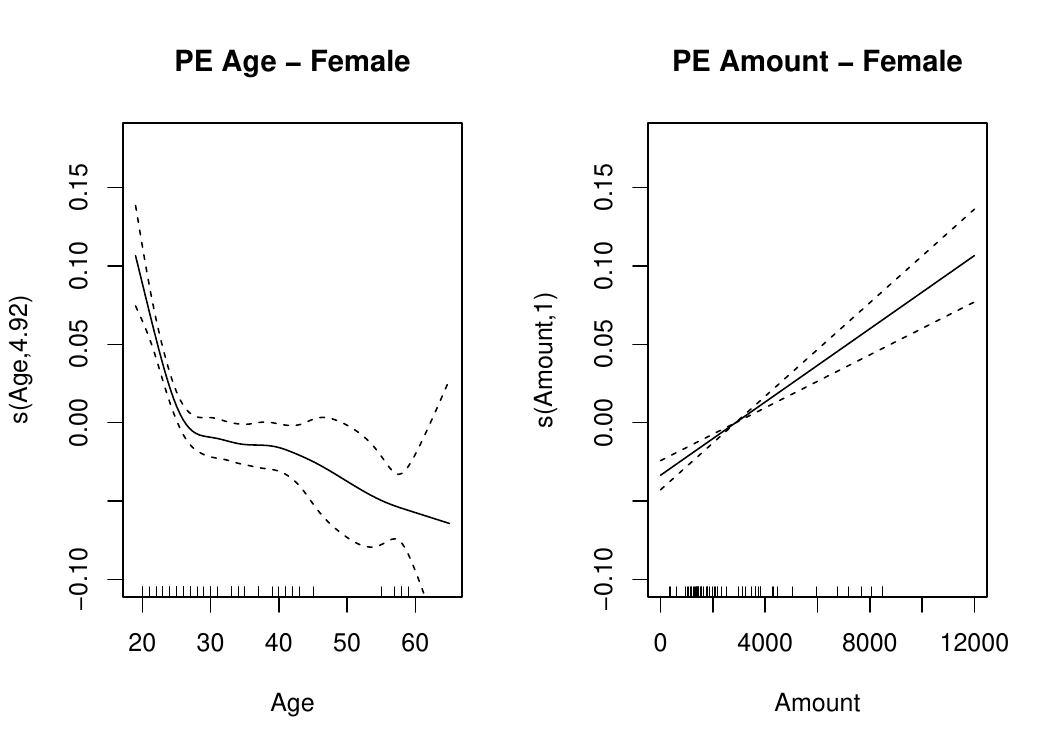}
         \caption{}
         \label{fig:german-1}
     \end{subfigure}
          \hfill
    \begin{subfigure}[b]{0.49\textwidth}
         \centering
         \includegraphics[width=\textwidth]{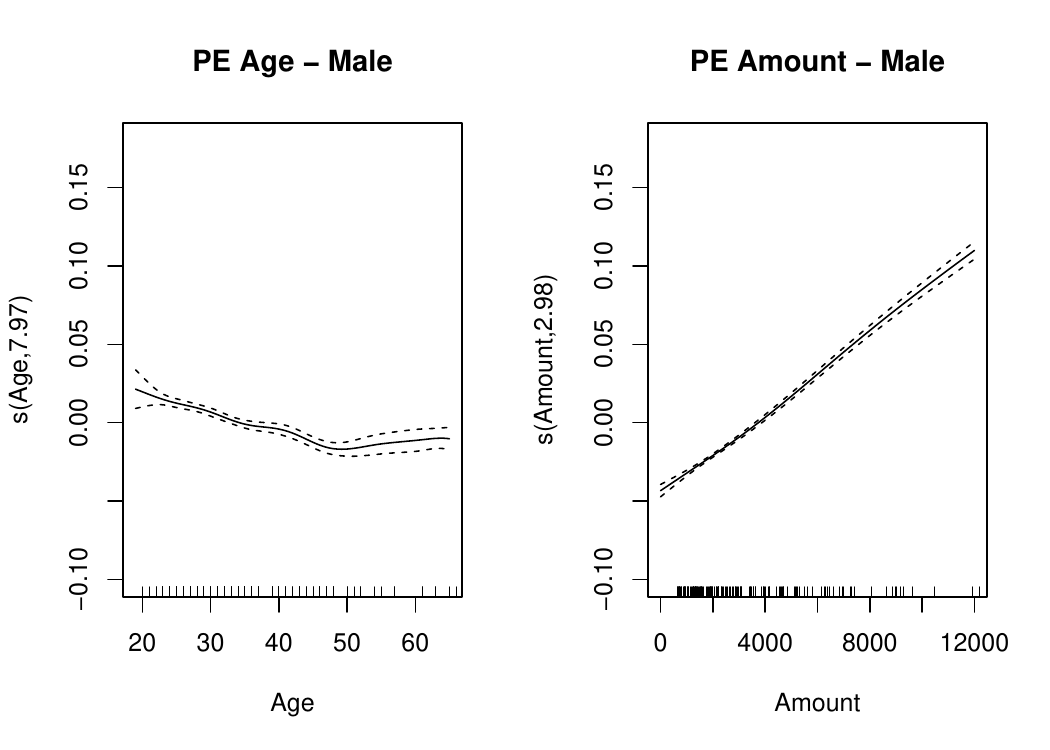}
         \caption{}
         \label{fig:german-3}
     \end{subfigure}
     \caption{Partial effect of Age and Amount on (a) female and (b) male prediction differences between warped and real world. Positive effects indicate higher values in warped world, i.e., negative discrimination in real world.}
\end{figure}

\paragraph{RPID:} Table \ref{tab:pred-diff-rpid} shows individuals whose predictions differ most in the two worlds, either positively or negatively. The observations above the dots have higher (i.e., better) warped-world predictions than real-world predictions, meaning they are negatively discriminated in the real world. The observations below the dots, on the other hand, are positively discriminated in the real world. Regressing these differences on features reveals that the risk prediction of young women grows strongly through warping, indicating that this subgroup was discriminated against most strongly in the real world (see Figure \ref{fig:german-1}). The partial effect of Amount is rather small in the area where most of the data live and seems only relevant for some very high values. 
The individual predictions for males change, too: Applying the same regression, Figure \ref{fig:german-3} shows partial effects of Age and Amount on the prediction difference. This reveals that young men also get higher predictions in the warped world -- but the effect is not as strong as for women.

\begin{figure}[ht]
    \centering
    \begin{subfigure}[b]{0.49\textwidth}
         \centering
         \includegraphics[width=\textwidth]{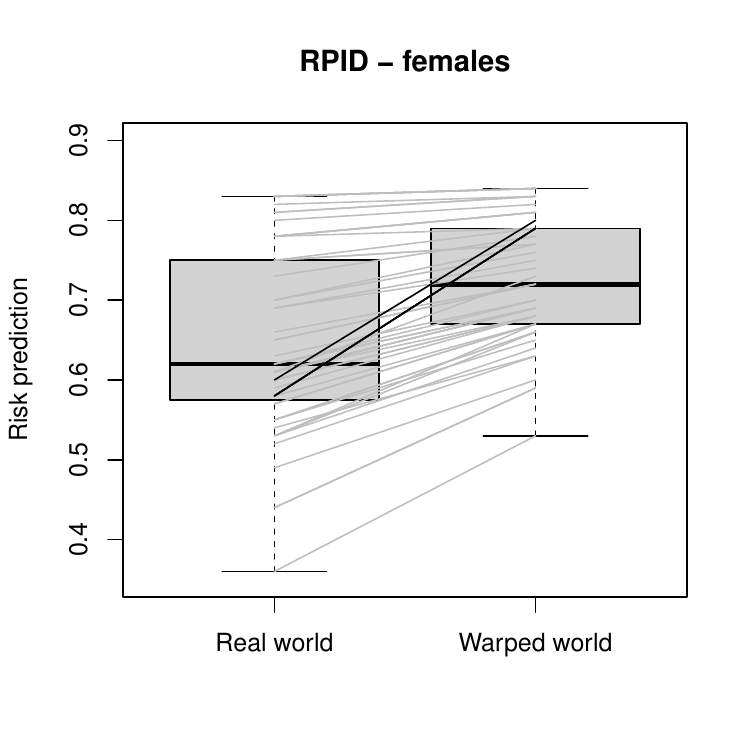}
         \caption{}
         \label{fig:german-4}
     \end{subfigure}
     \hfill
     \begin{subfigure}[b]{0.49\textwidth}
         \centering
         \includegraphics[width=\textwidth]{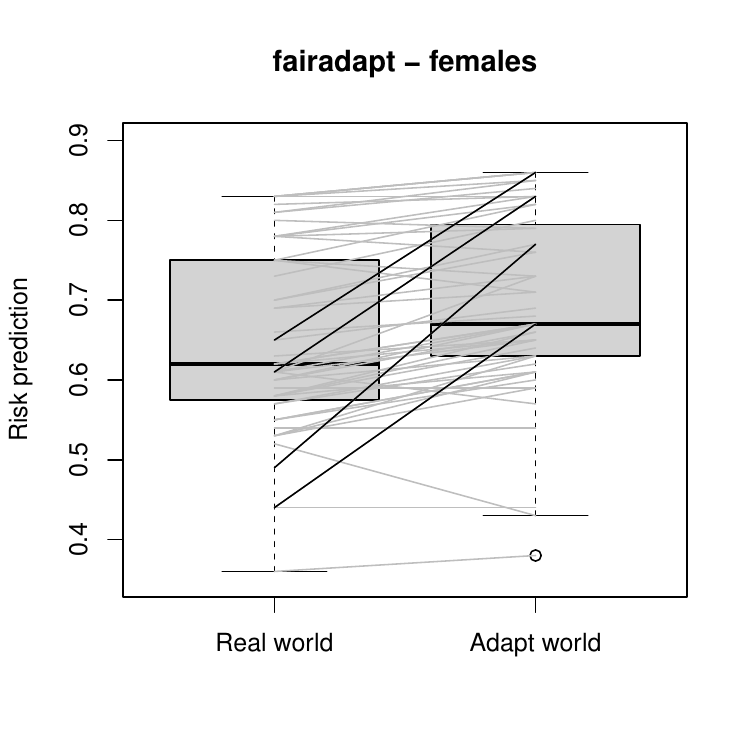}
         \caption{}
         \label{fig:german-4-adapt}
    \end{subfigure}
    \caption{Comparison of risk predictions for the female subgroup between (a) real and warped world and (b) real and adapt world. Grey lines connect same individuals, in each plot the four most strongly changed predictions are connected with black lines.}
    \label{fig:german}
\end{figure}

\begin{figure}[ht]
    \centering
    \begin{subfigure}[b]{0.49\textwidth}
         \centering
         \includegraphics[width=\textwidth]{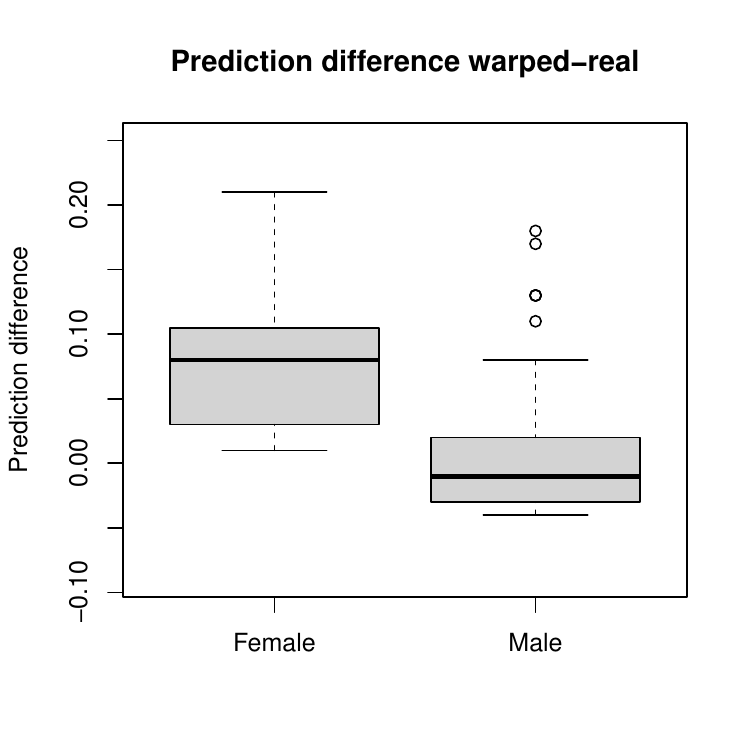}
         \caption{}
         \label{fig:german-2}
    \end{subfigure}
        \hfill
    \begin{subfigure}[b]{0.49\textwidth}
         \centering
         \includegraphics[width=\textwidth]{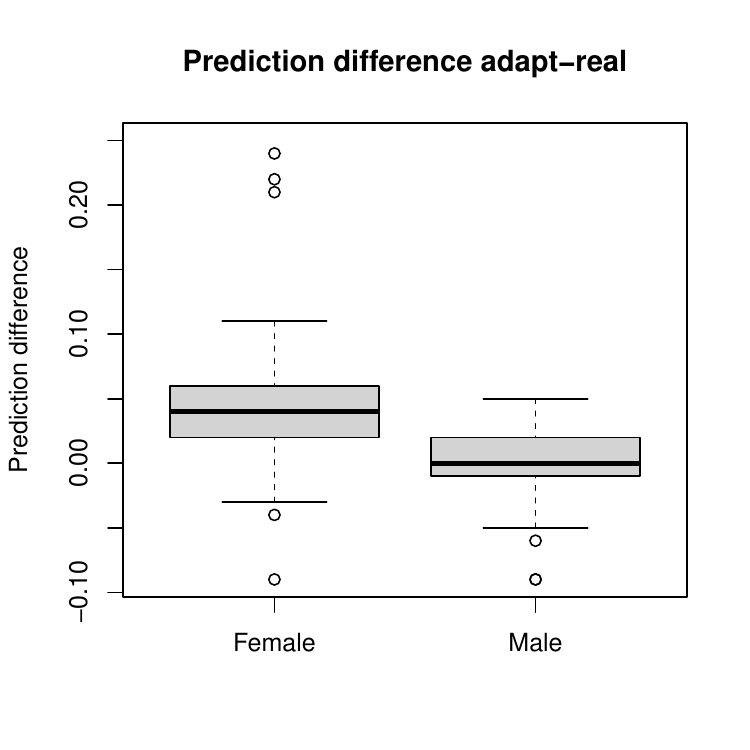}
         \caption{}
         \label{fig:german-2-adapt}
    \end{subfigure}
    \caption{Prediction differences between the two worlds for (a) RPID and (b) fairadapt.}
    \label{fig:german2}
\end{figure}

Figure \ref{fig:german-4} compares female predictions in both worlds and highlights the four most strongly affected individuals (the four individuals above the dots in Table \ref{tab:pred-diff-rpid}). This shows that all females get higher predictions in the warped world, but the internal ranking in the female subgroup changes, indicating that real-world discrimination is not equally high for all females and emphasizing that RPID is a quasi-individual approach. Figure \ref{fig:german-2} shows prediction differences for female and male subgroups. While mean differences for females are significantly positive ($p<10^{-12}$), male predictions do not change significantly on average ($p=0.32$). 

\paragraph{fairadapt:}
Table \ref{tab:pred-diff-adapt} shows individuals whose predictions differ most in the two worlds, either positively or negatively. The range is comparable to the results of RPID.
Figure \ref{fig:german-4-adapt} compares female predictions in both worlds and highlights the four most strongly affected individuals. Figure \ref{fig:german-2-adapt} shows prediction differences for female and male subgroups. While mean differences for females are significantly positive ($p<10^{-12}$), male predictions do not change significantly on average ($p=0.85$). 

\begin{table}%

  \centering
  \subfloat[][]{    
  \label{tab:pred-diff-rpid}
    \begin{tabular}{rrrrrrrrr}
    \toprule
    $Gender$ & $Age$ & $X_A$ & $X_S$ & $\hat{X}_A$ & $\hat{X}_S$ & $\pihi$ & $\pithi$ & \textit{diff}\\
    \midrule
    female & 22 &  1567 &     1 & 2095 & 0& 0.58 & 0.79 &   0.21 \\
    female &20 &  1282 &     1 & 1597 & 0 &   0.58 & 0.79 & 0.21 \\
    female & 22 &  1808 &     1 & 2227 & 0 &   0.58 & 0.79 & 0.21 \\
    female &21 &  1049 &     1 & 1188 & 0 &   0.60 & 0.80 & 0.20 \\
    \ldots &\ldots &  \ldots &     \ldots & \ldots &\ldots &   \ldots \\
    male &57  & 1264   &   0& 1264   &   0 &0.91     & 0.87& -0.04 \\
    male &66 &   766    &  0&   766    &  0    &0.93   &   0.89&  -0.04\\
    \bottomrule
    \end{tabular}
    }%
  \qquad
  \subfloat[][]{    
  \label{tab:pred-diff-adapt}
    \begin{tabular}{rrrrrrrrr}
    \toprule
    $Gender$ & $Age$ & $X_A$ & $X_S$ & $\hat{X}_A$ & $\hat{X}_S$ & $\pihi$ & $\pithi$ & \textit{diff}\\
    \midrule
female & 24 & 4308 & 1 & 4379 & 0 & 0.49 & 0.77 & 0.28 \\ 
  female & 37 & 7685 & 1 & 8489 & 0 & 0.44 & 0.67 & 0.24 \\ 
  female & 26 & 1453 & 1 & 1928 & 0 & 0.61 & 0.83 & 0.22 \\ 
  female & 27 & 343 & 1 & 338 & 0 & 0.65 & 0.86 & 0.21 \\ 
    \ldots &\ldots &  \ldots &     \ldots &\ldots &\ldots &    \ldots \\
  male & 74 & 4526 & 1& 4526 & 1 & 0.80 & 0.71 & -0.09 \\ 
  male & 68 & 14896 & 1& 14896 & 1 & 0.42 & 0.33 & -0.09 \\ 
    \bottomrule
    \end{tabular}
    }
    \caption{Most discriminated individuals for German Credit data, Table (a) for RPID, Table (b) for \textit{fairadapt}. Features $X_A$ and $X_S$ are warped to $\hat{X}_A$ and $\hat{X}_S$, respectively, $\pihi$ are predictions in the real world, $\pithi$ are predictions in the warped world or adapt world, respectively, and \textit{diff}$=\pithi-\pihi$.}
    \label{tab:pred-diff}
\end{table}

\subsubsection{Identifying strongest warped individuals -- UC3}

Investigating the effect of warping on the individuals reveals similar results as investigating the prediction differences in (UC2) and is omitted for the sake of concise presentation. %Table XXX shows the (female) individuals that are strongest affected by the warping. Regressing the distance between the two worlds on the features reveals that young women undergo the strongest warping. \lb{..or something like that, have to look at the results first.}

\subsubsection{Identifying important features -- UC4}

\paragraph{RPID:} The normalized feature differences between the real world and the warped world for Age, Amount, and Savings are $0.00, 0.01, 0.24$, respectively. This reveals that Savings is affected most by the warping and, hence, carries the strongest gender discrimination effect in the real world.

\paragraph{fairadapt:}
The normalized feature differences between the real world and the warped world for Age, Amount, and Savings are $0.00, 0.07, 0.28$, respectively. This reveals that as for RPID, Savings is affected most by the adaptation and, hence, supports the above finding that this feature carries the strongest gender discrimination effect in the real world.

%\newpage

\subsubsection{Discussion} 

The above analyses offer interesting insights: By computing the differences between real-world and warped-world risk predictions, most strongly discriminated individuals could be identified. Furthermore, these differences could be related to features such as Gender and Age, indicating that young females are discriminated against most strongly in the real world. In this sense, RPID results are comparable to those using \textit{fairadapt}.
This is interesting because our proposed residual-based approach uses rather simple generalized linear models instead of the more complex quantile regression forests used by \textit{fairadapt}.

Practical feasibility: The computational costs of our method are rather small for the presented analysis. The computations for the German credit data (learning warping models, training models in both worlds, warping, and predicting in the warped world) took $0.28$ seconds for RPID on a 3,4 GHz Intel Core i5, using one core of the CPU; computing time for \textit{fairadapt} was $1.25$ seconds, so roughly five times as much. This increases with, e.g., (i) the data size, (ii) the complexity of the DAG (since more models have to be trained), and (iii) the ML models used (e.g., training multilayered neural networks takes more time than training the logit models used here). We do not expect the computational time to be a relevant constraint -- also because learning the different models for warping can be parallelized.

\section{Conclusion and Outlook}
\label{sec:conclusion}

We have presented rank-preserving interventional distributions as a framework to identify a FiND world that can be considered to be fair because no causal effects of PAs exist; they are philosophically sound as they do not require a conceptualization of an intervention on the PAs and they aim for quasi-individual fairness in a rank-preserving sense. Additionally, we have proposed a warping method for estimating FiND world distributions with real-world data. A simulation study showed that the method works for the investigated simulation setup (see Section \ref{sec:discussion_sim} for limitations), and we demonstrated how the method can be applied to empirical data (\ref{sec:results_german}). Analyses can be reproduced via a public GitHub repository, which also contains code for applied use cases.\footnote{%Made available upon acceptance, to preserve anonymity.}
\url{https://github.com/slds-lmu/paper\_2023\_cfml}}
Apart from extending the study as outlined in Section \ref{sec:discussion_sim}, further work should compare our method to other methods that conceive a fictitious world for tackling fairness issues of ML models (see references in Section \ref{sec:related-work}).

While we have described our warping procedure for a specific DAG and assumed that identification assumptions are met, it can be used straightforwardly with more complex DAGs with more nodes. However, future work should engage with extensions for DAGs that are structurally more complex (instead of just larger). Specifically, the questions of continuous PAs, intersectionality in the PAs, and multiple time points are important future challenges.
 
Another challenge for practical application is the dependence on knowing the true DAG. 
%Further research should not only engage in developing new methods for deriving a DAG based on expert knowledge and empirical data but should 
Early current research tackles the quantification of uncertainty inherent to the entire process of first finding the true DAG and then carrying out causal inference \shortcite{chang_post-selection_2024}. 
Only by properly estimating the errors that can be attributed to the different steps of the pipeline (such as deriving the DAG, estimating the causal effects, estimating individual quantiles, estimating prediction models, and the non-reducible aleatoric uncertainty), reliable methods for trustworthy ML can be composed. We hope our work inspires other researchers to engage in these important topics of fairness-aware ML.

\section*{Ethical Statements}

\paragraph{Ethical Considerations Statement}

Our work has a rather theoretical, methodological perspective, and we do not propose a concrete application that could harm individuals. 
While conducting this work, we focused on an individual understanding of fairness and proposed algorithms to mitigate such unfairness issues in ML. We hope that we explained sufficiently why we took this perspective.

In the examples, we used the feature Gender as a protected attribute. For ease of presentation, we assumed this to be a binary variable, pointing out that generalizations to multi-categorical Gender are straightforward. This must not be misunderstood as a position in favor of a binary view of Gender.

\paragraph{Researcher Positionality Statement}

The author group has extensive experience 
%is educated and based 
in the fields of statistics, computer science, and causal inference. While working in methodological research of ML, they appreciate philosophical questions surrounding the analysis of data, being skeptical against technical proposals where authors seem not to think clearly about \textit{why} they are proposing these solutions. Specifically, the authors draw a greater awareness of philosophical, ethical, and legal questions from interdisciplinary cooperations with researchers from (legal) philosophy.

\paragraph{Adverse Impact Statement}

This is a first proposal of how to implement rank-preserving interventional distributions to mitigate unfairness issues in ML. As pointed out in the discussion, several limitations have to be addressed in future work. Overtrusting our proposed residual-based method might have an adverse impact if used without further considering the application at hand. 
A crucial point is finding the DAG in the real world: It is of paramount importance that it is carefully designed using expert knowledge and empirical evidence. If our method is used carelessly or even willfully wrong, this might lead to establishing undesirable, i.e., discriminatory behavior of an ADM system. Transparency on the normative stipulations underlying the use of our method is crucial to mitigate adverse impacts.

\begin{acks}
  We thank Holger Löwe for helping with visualizations.
\end{acks}

%\newpage
\appendix

\section{Classical FairML Metrics}
\label{sec:app_classical_fairml}

%As elaborated on thoroughly in \cite{anonymous_2024}, %bothmann_what_2023
``Classical`` fairML metrics such as demographic parity, equalized odds, etc., do not reflect a clearly defined concept of fairness and, hence, are not suitable for deciding if an ML model entails unfairness. Hence, they can also not be used as quality criteria for evaluating our warping method. However, since some of these metrics are still popular, one might be interested -- from a descriptive or explorative point of view -- how these metrics change after applying the above proposed warping approach. They can also be viewed as ``fairness-related performance metrics'', allowing for more nuanced insights into the predictive performance of a model. For these reasons, we provide respective results in the following, strongly emphasizing that such results are neither suitable for proving nor for disproving that our method works.

\paragraph{Simulation study.}

For the simulation study described above, Tables \ref{tab:classical-1} -- \ref{tab:classical-3} summarize some group fairness metrics \cite<see, e.g.,>[for an overview]{verma_fairness_2018,caton_fairness_2023}. We display ratios of different metrics, comparing the male and female subgroup, where a value smaller than $1$ indicates that the respective metric in the male subgroup is larger than in the female subgroup (i.e., female value divided by male value). The tables show the following mean values (averaged over simulation iterations)\footnote{We used the R Package \texttt{fairmodels} \cite{wisniewski_fairmodels_2022}.}:
\begin{itemize}
    \item \texttt{ACC:} Ratio of subgroup-specific accuracies, a.k.a. \textit{Overall accuracy equality}
    \item \texttt{PPV:} Ratio of subgroup-specific positive predictive values (precisions), a.k.a. \textit{Predictive parity}
    \item \texttt{FPR:} Ratio of subgroup-specific false positive rates, a.k.a. \textit{Predictive equality}
    \item \texttt{TPR:} Ratio of subgroup-specific true positive rates, a.k.a. \textit{Equal opportunity}
    \item \texttt{STP:} Ratio of subgroup-specific positively predicted rates, a.k.a. \textit{Statistical parity} or \textit{Demographic parity}
    \item \texttt{No.\ checks passed:} In each iteration, we check for each of the values of \texttt{ACC, PPV, FPR, TPR, STP} if it is inside the interval $(\epsilon, \frac{1}{\epsilon})$, where we use $\epsilon = 0.95$ as tolerance value. This number reports the total number of checks passed, which is $\in \{0, 1, 2, 3, 4, 5\}$ for each iteration, i.e., the mean (as reported in the tables) is $\in (0,5)$.
\end{itemize}

Table \ref{tab:classical-1} shows that for the scenario of correctly specified DAG and using the larger subgroup (\textit{male}) as a reference group, the metrics are considerably closer to $1$ for the warped and FiND world, compared with the real world. Warped and FiND world values are very close, only for FPR there seems to be a (small) difference between warped and FiND world values.

% latex table generated in R 4.2.1 by xtable 1.8-4 package
% Mon Sep  4 08:57:50 2023
\begin{table}[ht]
    \caption{Group fairness metrics w.r.t.\ RQ1, i.e., no misspecification and no reverse warping.}
    \label{tab:classical-1}
    \centering
\begin{tabular}{lrrrrrr}
    \toprule
World & ACC & PPV & FPR & TPR & STP & No. checks passed \\ 
    \midrule
Real & 0.9391 & 0.9337 & 1.0409 & 0.9563 & 0.8718 & 1.4440 \\ 
Warped & 1.0041 & 1.0023 & 0.9760 & 1.0028 & 1.0004 & 4.7040 \\
Adapt & 0.9817 & 0.9869 & 1.0892 & 0.9883 & 0.9889 & 4.2650 \\ 
FiND & 0.9998 & 0.9999 & 1.0019 & 0.9999 & 0.9997 & 4.6040 \\ 
    \bottomrule
\end{tabular}
\end{table}

Table \ref{tab:classical-2} shows that for the scenario of a misspecified DAG, warped-world values are slightly more different than FiND world values but still considerably closer to $1$ than real-world values. Only for FPR, real-world values are closer to $1$ than warped-world values.

\begin{table}[ht]
    \caption{Group fairness metrics w.r.t.\ RQ2, i.e., misspecification and no reverse warping.}
    \label{tab:classical-2}
    \centering
\begin{tabular}{lrrrrrr}
    \toprule
World & ACC & PPV & FPR & TPR & STP & No. checks passed \\ 
    \midrule
Real & 0.9101 & 0.9022 & 1.0751 & 0.9401 & 0.8178 & 0.4070 \\ 
Warped & 1.0350 & 1.0207 & 0.7103 & 1.0208 & 1.0000 & 4.0000 \\ 
Adapt & 0.9620 & 0.9680 & 1.0886 & 0.9773 & 0.9432 & 3.5900 \\ 
FiND & 1.0001 & 1.0001 & 0.9999 & 1.0000 & 0.9997 & 4.5180 \\ 
    \bottomrule
\end{tabular}
\end{table}

Table \ref{tab:classical-3} shows that for the scenario of reverse warping, the picture is somewhat inconclusive: ACC and TPR of the real and warped world are very similar, for FPR the real-world values are closer to $1$ where for PPV and STP the warped-world values are closer to $1$. In the warped world, $76\%$ more checks are passed compared to the real world. 

\begin{table}[ht]
    \caption{Group fairness metrics w.r.t.\ RQ3, i.e., no misspecification and reverse warping.}
    \label{tab:classical-3}
    \centering
\begin{tabular}{lrrrrrr}
    \toprule
World & ACC & PPV & FPR & TPR & STP & No. checks passed \\ 
    \midrule
Real & 0.9391 & 0.9337 & 1.0409 & 0.9563 & 0.8718 & 1.4440 \\ 
Warped & 0.9393 & 0.9544 & 1.4364 & 0.9556 & 1.0012 & 2.5320 \\ 
Adapt & 1.0170 & 1.0116 & 0.9263 & 1.0113 & 0.9950 & 4.2820 \\ 
FiND & 0.9999 & 1.0000 & 0.9992 & 0.9997 & 0.9992 & 4.6390 \\ 
    \bottomrule
\end{tabular}
\end{table}

%\newpage

\begin{figure}[ht]
     \centering
     \begin{subfigure}[b]{0.49\textwidth}
         \centering
         \includegraphics[width=\textwidth]{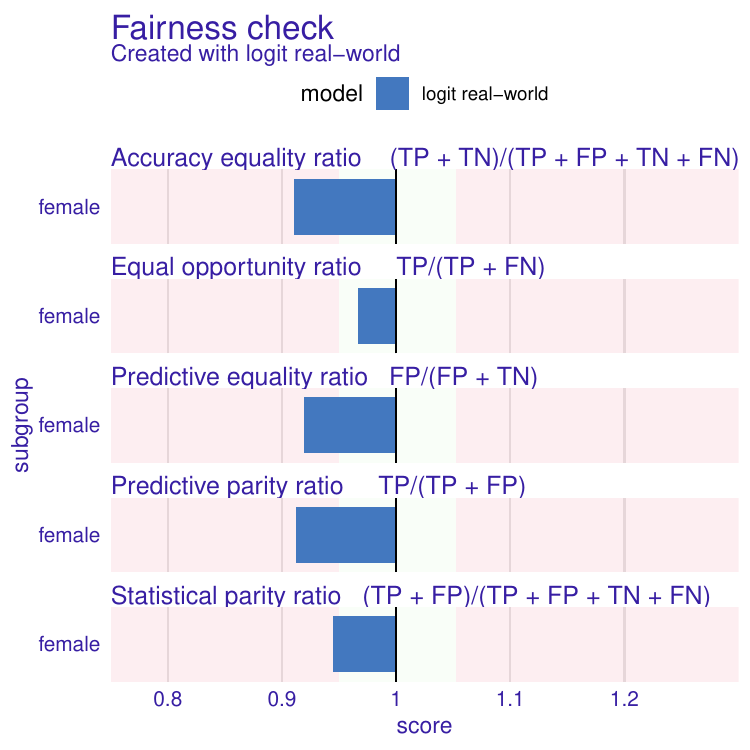}
         \caption{
         }
         \label{fig:german_fair_real}
     \end{subfigure}
     \hfill     
     \begin{subfigure}[b]{0.49\textwidth}
         \centering
         \includegraphics[width=\textwidth]{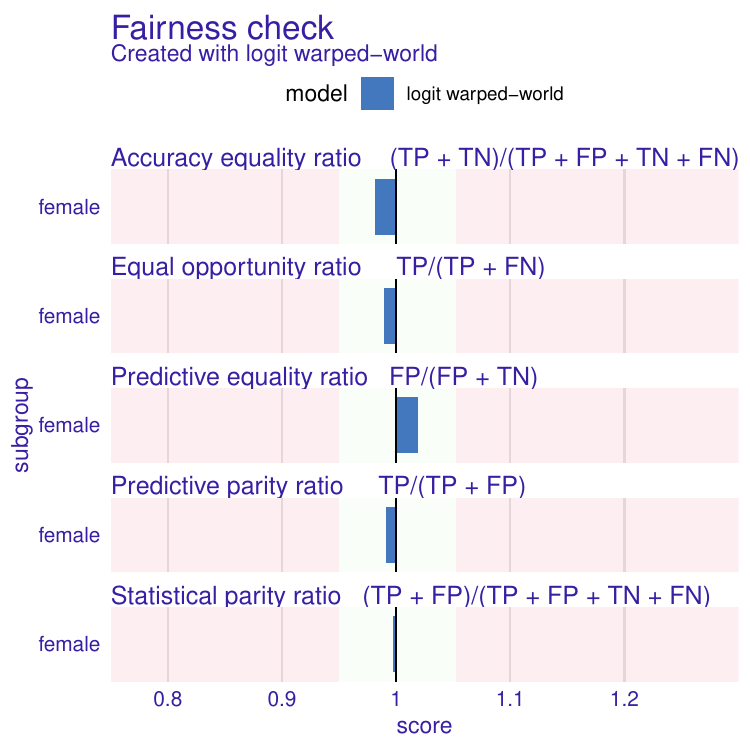}
         \caption{
         }
         \label{fig:german_fair_warped}
     \end{subfigure}\\
     \begin{subfigure}[b]{0.49\textwidth}
         \centering
         \includegraphics[width=\textwidth]{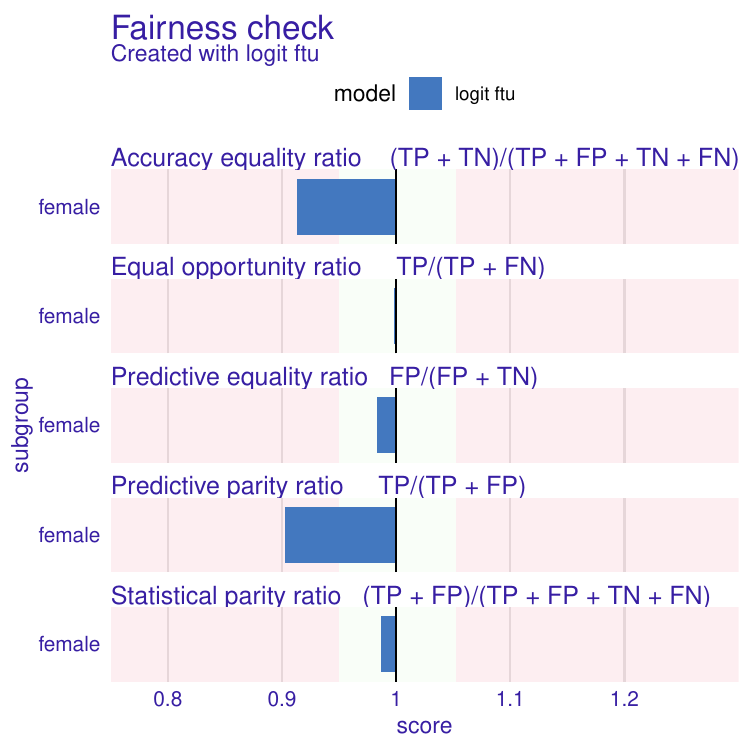}
         \caption{         }
         \label{fig:german_fair_ftu}
     \end{subfigure}
     \hfill     
     \begin{subfigure}[b]{0.49\textwidth}
         \centering
         \includegraphics[width=\textwidth]{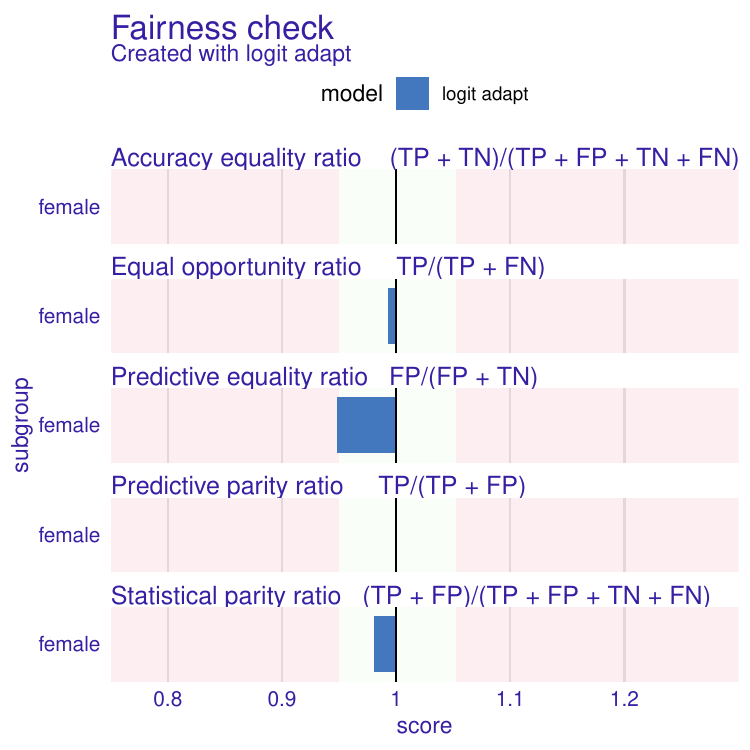}
         \caption{
         }
         \label{fig:german_fair_adapt}
     \end{subfigure}
        \caption{Group fairness metrics for German credit data in (a) real world, (b) warped world, (c) real world using FTU, and (d) adapt world, produced by R-Package \texttt{fairmodels} \cite{wisniewski_fairmodels_2022}.}
        \label{fig:german_fair}
\end{figure}

\paragraph{German Credit Data.} Figure \ref{fig:german_fair} shows the same metrics for the analysis of the German credit data. Figure \ref{fig:german_fair_real} shows metrics using all features in the real world and Figure \ref{fig:german_fair_ftu} compares this to a model excluding Gender, i.e., applying fairness through unawareness (FTU) on the same dataset. Figure \ref{fig:german_fair_warped} shows results for RPID, i.e., data in the warped world, and Figure \ref{fig:german_fair_adapt} compares this to a model trained on adapt-world data. For the full model in the real world, $1$ check is passed (equal opportunity ratio), for FTU, $3$ checks are passed, for RPID, all $5$ checks are passed, and for \textit{fairadapt}, $4$ checks are passed.

\newpage

%\bibliographystyle{ACM-Reference-Format}
%\bibliography{sample-base}
\bibliography{mybib_jair} % mybib is export from zotero, changed some smaller things to cope with theapa style
%\bibliography{mybib_anonym} % Anonymized version of fairness paper
\bibliographystyle{theapa}

\end{document}